%% file: main.tex
\documentclass[11pt]{article}

\usepackage[preprint]{acl}

\usepackage{times}
\usepackage{latexsym}

\usepackage[T1]{fontenc}
\usepackage[utf8]{inputenc}

\usepackage{microtype}

\usepackage{fontawesome5}

\usepackage{inconsolata}

\usepackage{graphicx}
\usepackage{amsmath}
\usepackage{amssymb}
\usepackage{xcolor}
\usepackage{listings}

\lstdefinestyle{promptbox}{
  basicstyle=\ttfamily\scriptsize,
  breaklines=true,
  breakatwhitespace=false,
  columns=fullflexible,
  keepspaces=true,
  frame=single,
  framerule=0.2pt,
  rulecolor=\color{black!25},
  backgroundcolor=\color{black!5},
  xleftmargin=2pt,
  xrightmargin=2pt,
  aboveskip=6pt,
  belowskip=6pt
}

\newcommand{\ours}{\textsc{ChemCoTBench-V2}}

\usepackage{subfiles}

\title{From Answers to States: Verifiable Process-Level Evaluation of Chemical Reasoning in Large Language Models}

\author{
 \textbf{Hongyu Guo\textsuperscript{1}},
 \textbf{Hao Li\textsuperscript{1, 2}},
 \textbf{He Cao\textsuperscript{2}},
 \textbf{Gongbo Zhang\textsuperscript{1}},
 \textbf{Li Yuan\textsuperscript{1 \footnotemark[1]}}
\\
 \textsuperscript{1} Peking University, Shenzhen Graduate School \\
 \textsuperscript{2} International Digital Economy Academy (IDEA)
\\
 \small{
   \textbf{Correspondence:} \href{mailto:yuanli-ece@pku.edu.cn}{yuanli-ece@pku.edu.cn}
 }
\\
\small{\faGithub\ \url{https://github.com/fresnellll/ChemCoTBench-V2}}
\small{
\raisebox{-0.2em}{\includegraphics[height=1.2em]{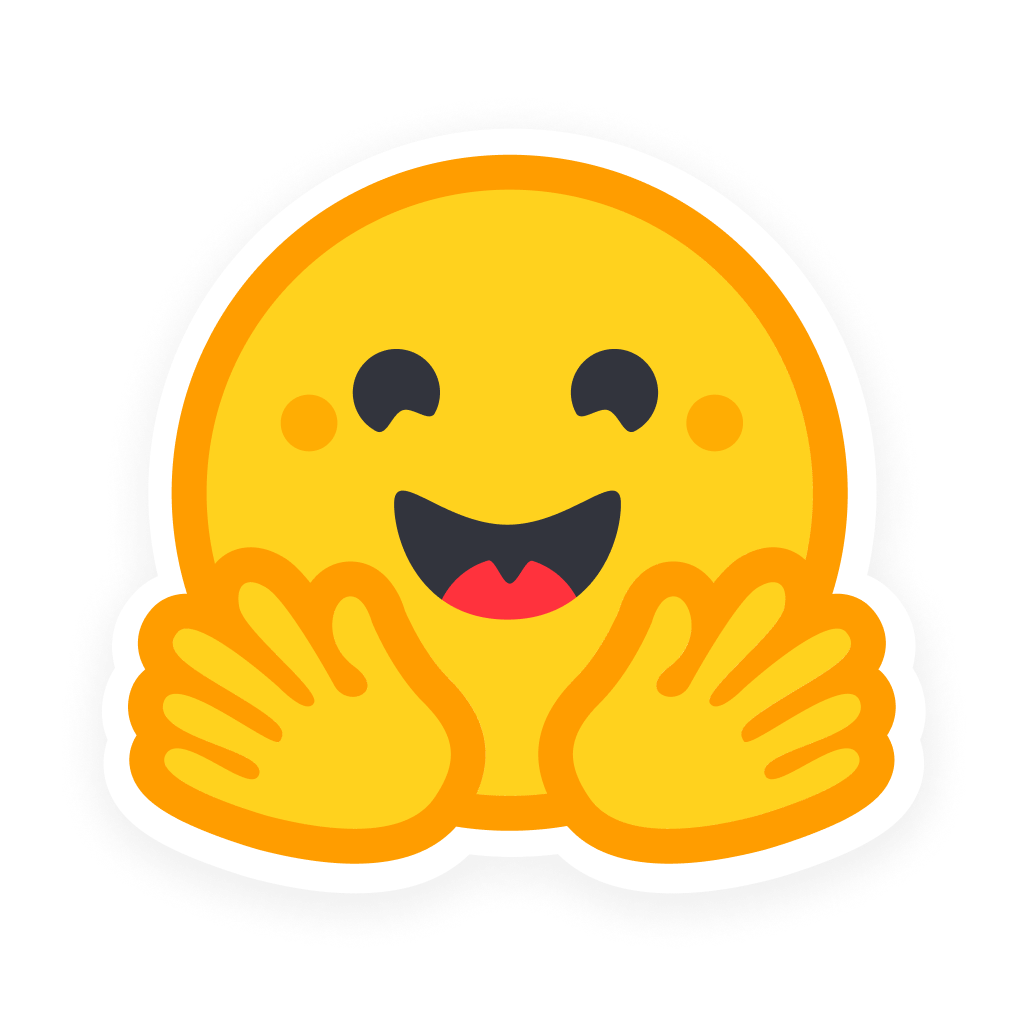}}
\ \href{https://huggingface.co/datasets/fresnellll/ChemCoTBench-V2}{ChemCoTBench-V2}
}
}

\begin{document}
\maketitle
\footnotetext[1]{Corresponding author.}
\begin{abstract}

Large language models are increasingly used as chemistry assistants, yet most chemistry benchmarks still score only final answers.
This masks a critical failure mode: a model may output the correct molecule, product, or option while its reasoning violates chemical logic.
Existing process-level evaluators are hard to scale because LLM judges and human step-level process annotation are costly, inconsistent, and vulnerable to hallucination.
We introduce \textsc{ChemCoTBench-V2}, a rule-verifiable diagnostic benchmark for low-cost, auditable evaluation of structured, verifier-addressable chemical reasoning traces.
It spans molecular understanding, molecule editing, molecular optimization, and reaction prediction, with 5,620 evaluation samples across 18 reporting tasks.
Models must expose key intermediate steps in expert-designed templates, and those steps are checked with deterministic chemistry rules and, for closed-answer tasks, reference traces rather than another LLM judge.
Open-ended molecular optimization is evaluated with oracle-verifiable state constraints rather than strict trace matching.
The benchmark reports three separate signals: final-answer correctness, template adherence, and step-wise verifier correctness over expert-refined intermediate commitments.
Experiments on frontier models reveal a persistent gap between final-answer success and structured-reasoning-state consistency: models often follow the requested format while failing chemical-step checks, or answer correctly with weak supporting reasoning.
\textsc{ChemCoTBench-V2} enables fine-grained model comparison and identifies the concrete step at which the trace first violates the verifier.

\end{abstract}

\input{Contents/1_Introduction}
\input{Contents/2_Relatedworks}
\input{Contents/3_Method}
\input{Contents/4_Exp}
\input{Contents/5_Conclusion}

\bibliography{custom}

\input{Contents/appendix}

\end{document}

%% file: Contents/1_Introduction.tex
\section{Introduction}

Large language models (LLMs) are becoming important tools for chemistry, from molecular question answering to reaction understanding~\cite{cao2023instructmol, zhang2024chemllm, chemdfm, pei2023biot5, chembench}.
The rise of reasoning-oriented models has made step-by-step chain-of-thought (CoT) a common interface for solving complex scientific problems~\cite{ether0, cao2026chemcotbench, Chemdfmr, luo2024improve}.
However, most chemistry benchmarks still evaluate these models as question-answering systems: they score the final SMILES string, option, ranking, or property value, while leaving the reasoning process untested~\cite{chembench, castro2023large, fang2024molinstructions, li2024speak, lu2024moleculeqa, huang2024chemeval}.
\begin{figure*}[t]
    \centering
    \includegraphics[width=\textwidth]{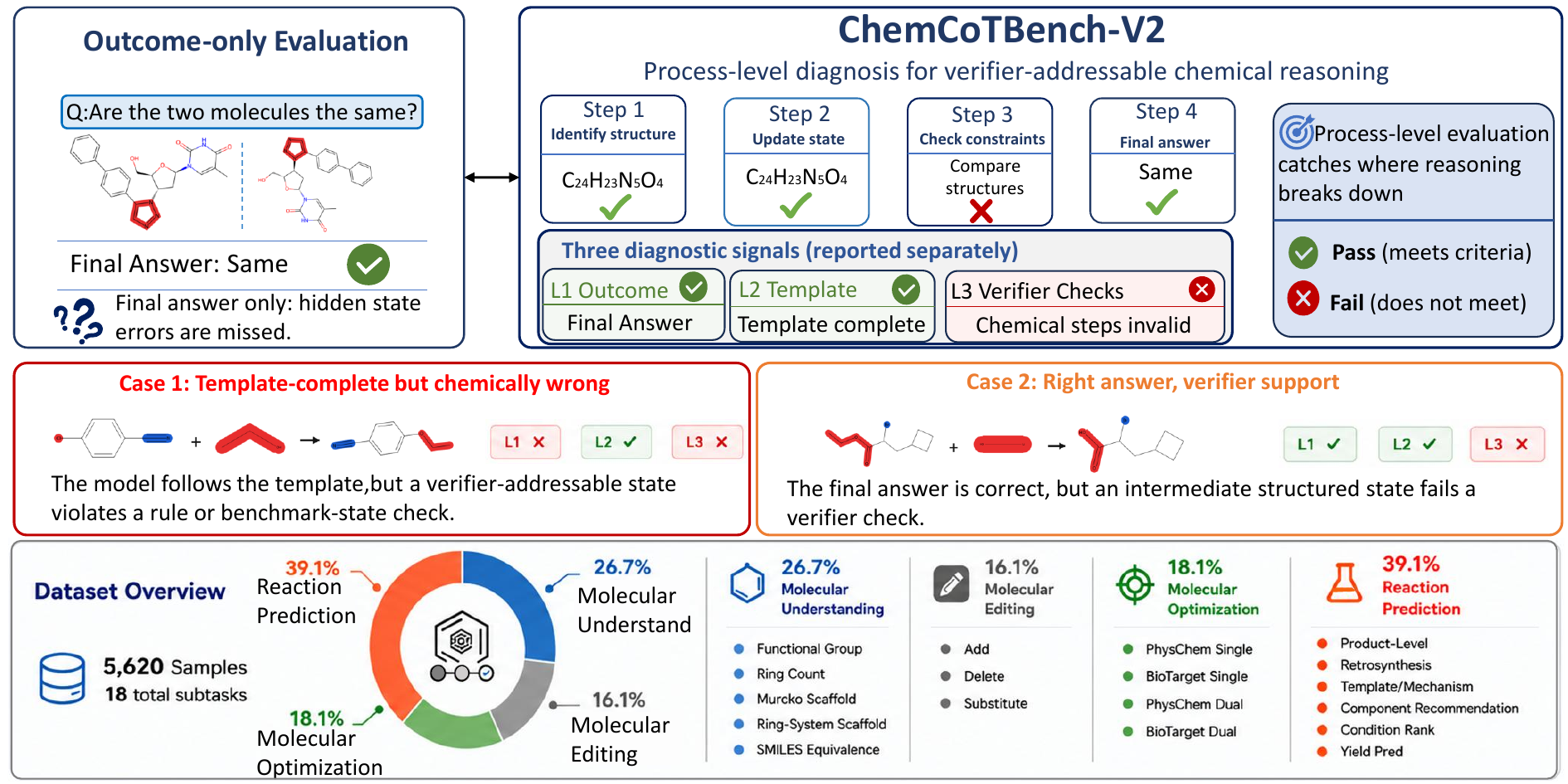}
    \vspace{-0.6cm}
    \caption{\ours{} evaluates structured, verifier-addressable chemical reasoning traces beyond final answers with three signals: Layer 1 outcome correctness, Layer 2 template adherence, and Layer 3 step-wise validity under deterministic task-specific checks.}
    \vspace{-0.4cm}
    \label{fig:intro_diagnosis}
\end{figure*}
This outcome-only view hides a critical failure mode.
A model may reach the correct final molecule or product while making an impossible bond change, misidentifying a scaffold, or violating a basic conservation rule along the way.
Without process-level evaluation, we can know that a model is wrong, but not where its verifier-addressable chemical commitment becomes inconsistent.

Process-level evaluation is therefore essential, but existing approaches are not yet practical for chemistry.
Human step-level annotation is costly, LLM-as-a-judge evaluation~\cite{li2024llms} is unreliable for fluent but chemically invalid explanations, and current rule-verifiable evaluations rarely cover dynamic tasks such as molecule editing, optimization, and reaction prediction.

We introduce \ours, a rule-verifiable diagnostic benchmark for structured chemical reasoning.
The key idea is to distill naturally occurring chemical CoT patterns into expert-refined, verifier-addressable intermediate commitments, and then evaluate whether models can maintain those commitments consistently across a formal reasoning trace.
It contains 5,620 active evaluation samples across 18 reporting tasks in four task families: molecular understanding, molecule editing, molecular optimization, and reaction prediction.
For closed-answer tasks, verified references define benchmark states for Type-II checking; for open-ended optimization, Layer 3 uses oracle-computable constraints.
We report three signals---outcome correctness, template adherence, and step-wise verifier correctness---so final-answer success can be separated from structured reasoning-state consistency.
We do not treat Type-II as exhaustive proof of all possible chemical rationales; it is benchmark-state agreement for tasks with deterministic or closed-form intermediate targets.

Experiments on frontier models show that final-answer success and structured chemical reasoning remain separable abilities.
Models can follow the requested template almost perfectly while failing step-wise verifier checks, and they can sometimes answer correctly with weak supporting reasoning.
Our contributions are: (i) a benchmark construction pipeline that distills natural chemical CoT traces into expert-refined, rule-verifiable intermediate commitments; (ii) a unified three-layer framework that separates outcome accuracy, template adherence, and step-wise verifier correctness across 5,620 samples and 18 reporting tasks; and (iii) a fine-grained diagnostic protocol that localizes verifier-detected reasoning-state inconsistencies through structured formal traces.

%% file: Contents/2_Relatedworks.tex
\section{Related Work}

\paragraph{Chemical reasoning benchmarks.}
Recent chemistry benchmarks evaluate LLMs on molecular questions, multimodal scientific tasks, and reaction understanding~\citep{fang2024molinstructions, wen2026innovatorvl, li2026rxnbench, zhao2025superchem, huang2024chemeval}.
These benchmarks have expanded the scope of chemical evaluation beyond simple property prediction, and some work further decomposes chemistry problems into modular operations such as structure recognition, molecule editing, optimization, and reaction prediction~\citep{cao2026chemcotbench, chembench}.
However, their reported metrics remain primarily outcome-based and do not directly verify intermediate chemical operations.
\ours{} targets this missing process-level dimension.

\paragraph{Process-level evaluation and rule verification.}
Outcome accuracy can overestimate model ability when correct answers are supported by invalid reasoning~\citep{wang2026outcomeaccuracy, shao2025deepseekmathv2}.
Existing process-level evaluation often relies on human labels, reward models, or LLM judges~\citep{lightman2023let, guan2025rstar, yuan2024free, zhang2024entropy, jacovi2024weakestlink, son2024llm}, which are costly or unreliable for chemistry.
In contrast, many chemical intermediate states are deterministically checkable once made explicit.
Prior work has shown the promise of symbolic verification on molecular graphs~\citep{bartmann2026moleculariq, runcie2025assessing, guo2024can}, but mainly for static reasoning rather than dynamic tasks such as editing, optimization, condition ranking, and reaction prediction.
Our work fills this gap with structured traces and rule-verifiable evaluation for verifier-addressable chemical reasoning commitments.

%% file: Contents/3_Method.tex
\section{Method: \ours}
\label{sec:Method}

\ours{} operationalizes process-level chemical reasoning as expert-refined, verifier-addressable intermediate commitments.
Instead of judging arbitrary free-form CoT sentence by sentence, it parses model responses into structured traces and tests whether their key chemical commitments are complete, internally consistent, and chemically verifiable.

\subsection{Task Taxonomy and Rule-Verifiable Instances}
\label{sec:task_taxonomy}

\ours{} covers 18 reporting tasks organized under four task families.
The molecular understanding tasks focus on structure perception, including functional groups, rings, scaffolds, and SMILES equivalence; the molecule editing tasks cover site-specific add, delete, and substitute operations; the molecular optimization tasks separate physicochemical and bioactivity optimization under single- and dual-objective settings; and the reaction prediction tasks span product-level prediction, retrosynthesis, template and mechanism reasoning, component recommendation, condition ranking, and yield prediction.

The instances are constructed so that both final answers and intermediate states are deterministically verifiable.
We draw molecules and reactions from public chemistry databases and task-specific pools, with full source details in Appendix~\ref{sec:appendix}.
RDKit-based sanitization and canonicalization remove invalid or ambiguous structures, strip answer-leaking metadata, and enforce task-specific constraints.
Molecule-editing examples are derived from real reactant--product changes and rewritten as site-specific edits, while condition-ranking labels are shuffled to prevent shortcut exploitation.
After filtering, redundancy reduction, and task-balanced sampling, the active evaluation set contains 5,620 rule-verifiable samples.

\subsection{Formal Reasoning Templates}
\label{sec:formal_templates}

Free-form chain-of-thought is difficult to audit reliably: a fluent explanation may mix correct chemistry with unsupported speculation and inconsistent formatting.
\ours{} therefore requires expert-designed templates that expose key intermediate chemical states.
Depending on the task, a template may specify fields such as SMARTS patterns, matched sites, reaction types, scaffold preservation, or product construction.
The same field names support both Layer~2 template-adherence checks and Layer~3 step verification.

\noindent\textbf{Template induction and expert refinement.}
Templates are induced before reference construction rather than derived from ground-truth-injected rationales.
We first collect natural direct-reasoning traces, use an LLM to summarize recurring reasoning fields, and then have chemistry experts remove fields that are not meaningful, stable, or deterministically verifiable.
This CoT-to-template distillation turns free-form reasoning into auditable commitments such as site identification, scaffold preservation, reaction-type selection, product construction, and constraint verification.

For instance $i$, we parse the response into a trace $\tau_i=\{s_{i,k}\}_{k=1}^{n_i}$, where each step is
$s_{i,k}=(z_{i,k}, r_{i,k}, f_{i,k})$.
Here, $z_{i,k}$ is the predefined step identifier, $r_{i,k}$ is a brief rationale, and $f_{i,k}$ is the structured output to be verified, such as a count, a SMILES/SMARTS string, a scaffold, a selected option, or a ranked list.
The model-facing format is:
\noindent\texttt{Step k [$z_{i,k}$]:} $r_{i,k}$ \textit{Structured output:} $f_{i,k}$.
This separation preserves readability while giving parsers and verifiers a stable target for extracting chemical states and decisions.

\begin{figure*}[t]
    \centering
    \includegraphics[width=\textwidth]{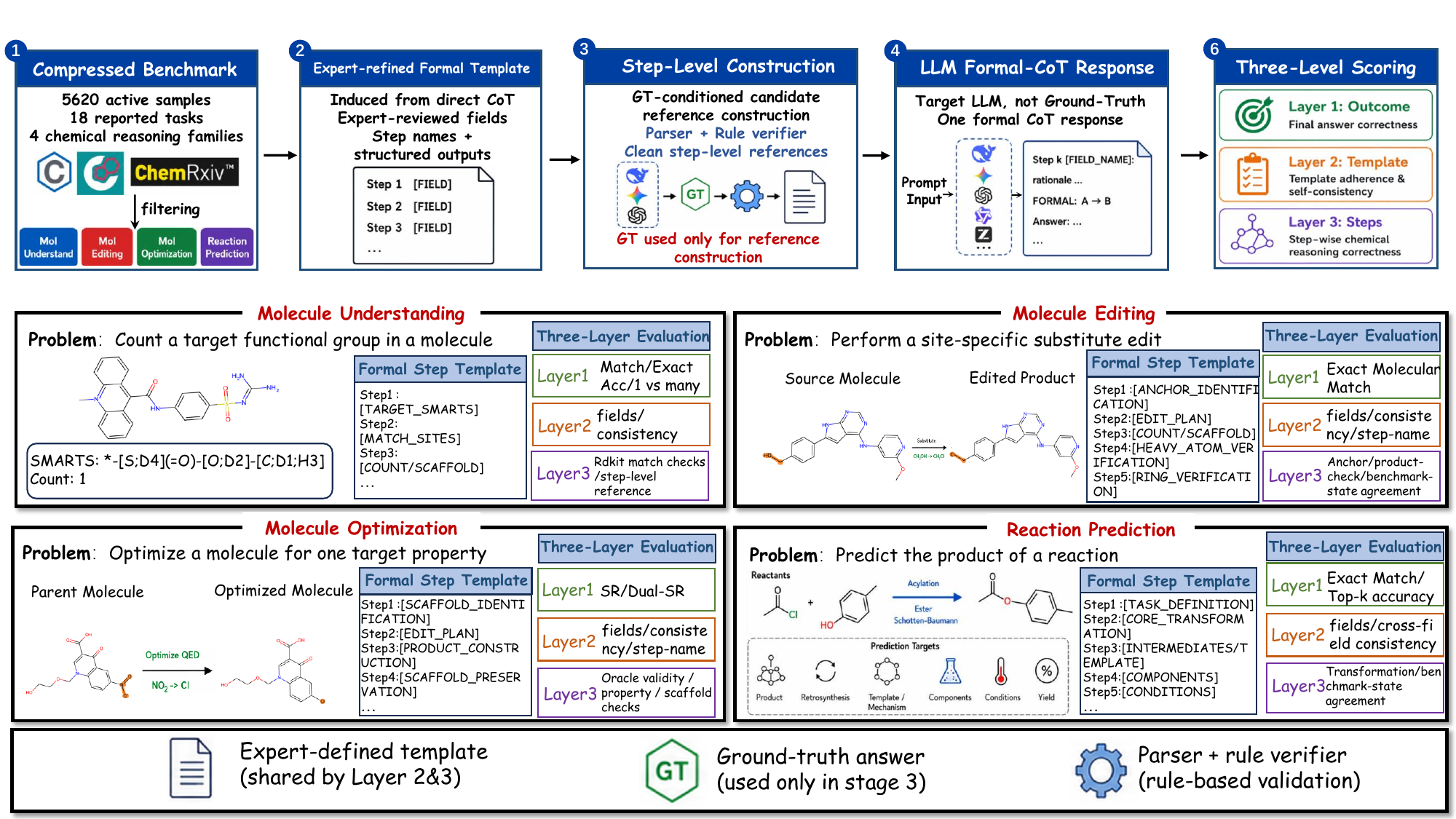}
    \vspace{-0.8cm}
    \caption{
    Unified framework for reference construction and evaluation.
% Step-level references are obtained by filtering GT-conditioned candidate traces with a parser and rule verifier. For evaluation, GT is removed from the prompt, and the target model is scored on a single formal-CoT response at the outcome, template, and step levels.
    }
    \vspace{-0.3cm}
    \label{fig:framework}
\end{figure*}

\subsection{Step-Level Reference Construction}
\label{sec:reference_construction}

After templates are fixed, GPT-5.4 and Claude-Opus-4.7 generate candidate references from the task input, ground-truth answer, and formal template.
Ground truth is used only in this reference-construction stage; evaluated models never see it.
Importantly, the template schema and evaluated reasoning fields are fixed beforehand, so GT conditioning instantiates candidate benchmark-state traces rather than defining the reasoning operations being evaluated.
Candidate traces are parsed and retained only if their required fields are extractable and they pass the corresponding Layer~1, Layer~2, and applicable Type-I checks.
Conflicts or unsafe repairs are manually inspected or excluded.
The verified references are used for Type-II benchmark-state agreement in closed-answer tasks; molecular optimization does not use them as strict path-matching targets.
A stratified expert audit of 300 traces is reported in Appendix~\ref{app:expert_validation}.

\subsection{Multi-Layer Diagnostic Protocol}
\label{sec:three_layer_metrics}

Let $N$ be the number of evaluated instances.
For the $i$-th instance, let $x_i$ be the parsed model response, $a_i$ its extracted final answer, and $y_i$ the ground-truth final answer.
For closed-answer Type-II checks, let $g_i$ denote the verified benchmark-state trace.
\ours{} reports three complementary layers.

\noindent\textbf{Layer 1: outcome correctness.}
Layer 1 evaluates only the final answer with task-appropriate metrics $M(a_i,y_i)$.
At the dataset level, we aggregate the outcome score as
{
\setlength{\abovedisplayskip}{1pt}
\setlength{\belowdisplayskip}{1pt}
\setlength{\abovedisplayshortskip}{1pt}
\setlength{\belowdisplayshortskip}{1pt}
\[
\mathrm{L1}=\frac{1}{N}\sum_{i=1}^{N} M(a_i,y_i),
\]
}
with the interpretation of $M$ determined by the task.
L1 uses standard task-specific outcome metrics: exact molecular match for editing; MAE, Tanimoto, or accuracy for molecular understanding; top-1 accuracy for categorical reaction tasks (with auxiliary ranking metrics for condition ranking), and MAE for yield prediction; and single- or dual-objective success rate for molecular optimization. This preserves comparability with conventional outcome-based chemistry benchmarks.

\noindent\textbf{Layer 2: template adherence.}
Layer 2 asks whether $x_i$ instantiates the requested scientific reasoning template.
It checks structural completeness, legal step names or enum values, presence of structured output fields, answer fields, and internal consistency between fields.
It deliberately excludes chemistry-tool execution and GT comparison, so a high Layer 2 score means the model followed the protocol, not that the chemistry is correct.
Let $\mathcal{V}_i=\{v_{i,1},\ldots,v_{i,m_i}\}$ be the task-specific set of Layer-2 template checks for instance $i$.
Each $v_{i,j}$ is a binary rule, such as checking that a required step is present, that an enum value is legal, or that a predicted value agrees with the model's own answer field.
We report the fraction of these template checks that pass:
$
\mathrm{StateScore}(x_i)=\frac{1}{m_i}\sum_{j=1}^{m_i}
\mathbb{I}\left[v_{i,j}(x_i)\right],
$
% where $\mathbb{I}[\cdot]$ is 1 when the logical condition or rule check is true and 0 otherwise.
The dataset-level Layer-2 score is then
{
\setlength{\abovedisplayskip}{1pt}
\setlength{\belowdisplayskip}{1pt}
\setlength{\abovedisplayshortskip}{1pt}
\setlength{\belowdisplayshortskip}{1pt}
\[
\mathrm{L2}=\frac{1}{N}\sum_{i=1}^{N}\mathrm{StateScore}(x_i).
\]
}
Type-II is benchmark-state agreement for closed-answer tasks, not exhaustive validation of all possible chemical rationales.

\begin{figure*}[!ht]
\centering
\scriptsize
\setlength{\fboxsep}{5pt}
\fcolorbox{gray!45}{gray!8}{
\begin{minipage}{\textwidth}
\textbf{Real diagnostic case: localizing a hidden reaction-type error.}
\hfill
\fcolorbox{green!45!black}{green!10}{\textcolor{green!45!black}{\textbf{correct/pass}}}
\quad
\fcolorbox{red!55!black}{red!10}{\textcolor{red!55!black}{\textbf{wrong/fail}}}

\vspace{2pt}
\textbf{Task.} Forward reaction prediction
(\texttt{pool\_id=3cabf5f0-fd6b-4063-b432-03a616b363e6}).

\textbf{Input.} \texttt{CCOC(=O)C(Br)CC1CCC1.[Li]O}

\textbf{Key prompt constraint.} The evaluated model is explicitly asked to use
the unified formal format from Section~\ref{sec:formal_templates}:
\texttt{Step k [$z_k$]:} rationale $r_k$;
\texttt{FORMAL: <input> -->} field $f_k$.
For \texttt{Step 2 [RXN\_TYPE]}, the prompt states:
\textit{``Select ONE from the 9 coarse-grained categories''}:
\texttt{C-C Coupling}, \texttt{Heteroatom Alkylation and Arylation},
\texttt{Acylation}, \texttt{Functional Group Interconversion},
\texttt{Deprotection}, \texttt{Reduction}, \texttt{Oxidation},
\texttt{Aromatic Heterocycle Formation}, and \texttt{Protection}.

\textbf{Condensed parsed trace.}\par\smallskip\noindent
{\setlength{\tabcolsep}{2pt}
\begin{tabular}{@{}p{0.04\linewidth}p{0.19\linewidth}p{0.30\linewidth}p{0.30\linewidth}p{0.07\linewidth}@{}}
$k$ & $z_k$ step name & $r_k$ short rationale & $f_k$ structured field & verifier \\
\hline
1 & \texttt{FG\_IDENTIFICATION} &
ester substrate; bromide retained; hydroxide reagent &
\texttt{FG\_LIST(["ester","bromide"])} &
\textcolor{gray!70}{--} \\
2 & \texttt{RXN\_TYPE} &
ester hydrolysis converts ester to carboxylic acid &
\texttt{RXN\_TYPE:} \textcolor{red!55!black}{Functional Group Interconversion} &
\textcolor{red!55!black}{\textbf{fail}} \\
3 & \texttt{MECHANISM} &
hydroxide attacks the carbonyl; alkoxide leaves &
\texttt{MECHANISM\_KWORD("hydrolysis")} &
\textcolor{gray!70}{--} \\
4 & \texttt{PRODUCT\_PREDICTION} &
ethyl ester is hydrolyzed; alkyl bromide remains &
\texttt{PRED\_SMILES:} \textcolor{green!45!black}{\texttt{OC(=O)C(Br)CC1CCC1}} &
\textcolor{green!45!black}{\textbf{pass}} \\
\end{tabular}
}
\par\smallskip

\textbf{Layer 1.}
\fcolorbox{green!45!black}{green!10}{\textcolor{green!45!black}{\textbf{pass}}}
Final product is correct:
\texttt{OC(=O)C(Br)CC1CCC1} is equivalent to the reference
\texttt{O=C(O)C(Br)CC1CCC1}.

\textbf{Layer 2.}
\fcolorbox{green!45!black}{green!10}{\textcolor{green!45!black}{\textbf{pass}}}
The trace is template-complete and internally consistent
(\texttt{State Score=1.0}): the required step names, the 9-way reaction-type
choice field, and the final answer field are present and self-consistent.

\textbf{Layer 3 localization.}
\fcolorbox{red!55!black}{red!10}{\textcolor{red!55!black}{\textbf{localized fail}}}
The product field passes
(\textcolor{green!45!black}{\texttt{gt\_match\_step4\_predicted\_smi=True}}), but the 9-way
\texttt{RXN\_TYPE} choice fails:

\begin{center}
\textcolor{red!55!black}{\texttt{model: choice 4, Functional Group Interconversion}}
$\neq$
\textcolor{green!45!black}{\texttt{reference: choice 5, Deprotection}}
\quad
(\texttt{fine label: CO2H-Et deprotection}).
\end{center}

The final answer is therefore right, but the trace violates the
benchmark-defined \texttt{RXN\_TYPE} reasoning state.
\end{minipage}}
\vspace{-0.25cm}
\caption{Sample-level diagnostic case from the Qwen3.5 Plus forward-reaction evaluation. The figure shows the prompt constraint, a shortened $(z_k,r_k,f_k)$ trace, and the three-layer diagnosis. Although the model predicts the correct product, its Step-2 reaction-type commitment disagrees with the benchmark-defined \texttt{RXN\_TYPE} state, which the verifier localizes as the failure.}
\label{fig:diagnostic_case}
\vspace{-0.3cm}
\end{figure*}

\noindent\textbf{Layer 3: step-wise verifier correctness.}
Layer 3 evaluates the verifier-addressable contents of the filled template.

\noindent\textit{Type-I intrinsic symbolic checks.}
Type-I predicates are computed without a reference trace.
They include SMILES validity, SMARTS matching, canonicalization, ring counting, heavy-atom arithmetic, scaffold containment, charge balance, or atom-conservation constraints.
Let $\mathcal{R}^{I}_i$ be the Type-I rule set selected by the task template for instance $i$.
The Type-I all-pass rate is
{
\setlength{\abovedisplayskip}{1pt}
\setlength{\belowdisplayskip}{1pt}
\setlength{\abovedisplayshortskip}{1pt}
\setlength{\belowdisplayshortskip}{1pt}
\[
\mathrm{L3}_{I} = \frac{1}{N}\sum_{i=1}^{N}
\mathbb{I}\left[\forall r\in\mathcal{R}^{I}_i,\ r(x_i)\right].
\]
}

\noindent\textit{Type-II benchmark-state agreement for closed-answer tasks.}
Type-II predicates are used for closed-answer intermediate states in molecular understanding, molecule editing, and reaction prediction.
They compare parsed fields with the verified benchmark-state trace $g_i$, such as scaffold, reaction class, product, ranked condition, selected option, or recommended component.
Let $\mathcal{R}^{II}_i$ be the corresponding Type-II benchmark-state agreement rule set.
The Type-II all-pass rate is
{
\setlength{\abovedisplayskip}{1pt}
\setlength{\belowdisplayskip}{1pt}
\setlength{\abovedisplayshortskip}{1pt}
\setlength{\belowdisplayshortskip}{1pt}
\[
\mathrm{L3}_{II} = \frac{1}{N}\sum_{i=1}^{N}
\mathbb{I}\left[\forall r\in\mathcal{R}^{II}_i,\ r(x_i,g_i)\right].
\]
}
Equivalently, these indicators implement a logical AND over the relevant step checks.
This strict criterion is intentional for tasks with tightly coupled symbolic logic: one invalid molecule, count, scaffold relation, reaction class, or final decision is enough to invalidate the corresponding verifier component.

\noindent\textit{Oracle-verified optimization L3.}
Molecular optimization is open-ended and is not evaluated with Type-II trace matching.
We instead use an oracle-verified optimization state score over a fixed-length optimization template.
Let $\mathcal{O}_i=\{o_{i,1},\ldots,o_{i,K}\}$ be the oracle predicates for instance $i$, covering validity, objective satisfaction, scaffold consistency, and consistency between the declared edit and generated molecule.
The optimization Layer-3 score is
{
\setlength{\abovedisplayskip}{1pt}
\setlength{\belowdisplayskip}{1pt}
\setlength{\abovedisplayshortskip}{1pt}
\setlength{\belowdisplayshortskip}{1pt}
\[
\mathrm{L3}_{\mathrm{opt}}=\frac{1}{N}\sum_{i=1}^{N}
\frac{1}{K}\sum_{k=1}^{K}\mathbb{I}\left[o_{i,k}(x_i)\right].
\]
}
In our implementation, $K=5$.
As elsewhere, this score evaluates the structured reasoning commitments exposed by the template fields, rather than unrestricted natural-language rationale text.

%% file: Contents/4_Exp.tex
\section{Experiments}

\subsection{Setup}
\label{sec:exp_setup}

We evaluate 8 frontier LLMs, covering both reasoning-oriented and standard instruction-following models, on the \ours{} benchmark.
All models use the same formal reasoning prompts, parsers, and rule-based verifiers.
The model suite is intentionally mixed: reasoning-oriented systems test whether explicit deliberation improves formal chemical traces, while standard instruction-following systems show how much of the benchmark can be solved by general chemical pattern recognition.
Section~\ref{sec:three_layer_metrics} defines the three layers at the instance level; here we report their aggregate statistics over each paper-facing task group.
Because the four task families use heterogeneous outcome metrics, we do not collapse Layer~1 into a single global score; instead, each table keeps the natural task metric and uses Layer~2/Layer~3 to compare trace quality.
Layer~1 (L1) uses task-specific outcome metrics, including accuracy, MAE, Tanimoto similarity, and success rate.
For molecular optimization, \textit{SR} denotes single-objective success rate, i.e., the percentage of generated molecules satisfying the target property-improvement criterion, and \textit{D-SR} denotes the percentage satisfying both target objectives simultaneously.
Layer~2 (L2) is the average State Score for template adherence and internal consistency.
Layer~3 (L3) measures step-wise verifier correctness: molecular understanding, molecule editing, and reaction prediction use Type-I all-pass and Type-II all-match rates, while molecular optimization uses an oracle-verified optimization state score.
Together, these statistics separate whether a model gets the answer right, follows the expected reasoning template, and maintains verifier-addressable chemical commitments.
When a trace fails, the structured fields further localize the error to a named chemical operation, such as scaffold extraction, product construction, reaction-type selection, condition ranking, or scaffold-preservation verification, rather than only marking the final answer as wrong.

\subsection{Task-Specific Reasoning Evaluation}

Before aggregating these failures by task family, Figure~\ref{fig:diagnostic_case} shows how the diagnostic protocol works at the sample level.
The model gives a correct final product and satisfies the template checks, but the formal trace exposes a wrong \texttt{RXN\_TYPE} commitment.
This is the same evidence pattern used throughout the task-specific analysis below: the tables report outcome and step scores, while the checkpoint logs identify the exact named operation where a reasoning trace breaks.

% 分子理解部分
% 写出反差：点出模型在fg_detect等偏向“模式匹配/文本层面”的任务上表现尚可，但是在ring_count (环计数) 和 murcko_scaffold (核心骨架提取) 上 Layer 3 严重掉点
% 分析原因：说明LLM基于SMILES的一维字符串表示缺乏2D/3D的拓扑图直觉。能够在 Layer 1 蒙对答案，但在 Layer 3 符号验证（重构骨架的每一步）时，原形毕露
\paragraph{Molecular understanding.}
Table~\ref{tab:molund} summarizes the five molecular understanding task groups.
These tasks expose a contrast between local SMILES-pattern recognition and 2D graph-topology reconstruction.
Models are relatively robust when the required commitment can be recovered from local string or substructure cues, but degrade when the trace must maintain an explicit graph object such as a ring set or Murcko scaffold.
This is clearest in the gap between SMILES equivalence and scaffold-centric tasks: several models reach strong outcome accuracy, yet their Layer-3 scores remain low because the intermediate scaffold or ring commitments are not chemically consistent.

\begin{table*}[t!]
\centering
\scriptsize
\setlength{\tabcolsep}{3.2pt}
\resizebox{\textwidth}{!}{
\begin{tabular}{l cc cc cc cc cc}
\hline
\textbf{Model} &
\multicolumn{2}{c}{\textbf{Functional Group}} &
\multicolumn{2}{c}{\textbf{Ring Count}} &
\multicolumn{2}{c}{\textbf{Murcko Scaffold}} &
\multicolumn{2}{c}{\textbf{Ring-System Scaffold}} &
\multicolumn{2}{c}{\textbf{SMILES Equivalence}} \\
& L1 MAE$\downarrow$ & L3 I/II$\uparrow$
& L1 MAE$\downarrow$ & L3 I/II$\uparrow$
& L1 Tan.$\uparrow$ & L3 I/II$\uparrow$
& L1 Acc.$\uparrow$ & L3 I/II$\uparrow$
& L1 Acc.$\uparrow$ & L3 I/II$\uparrow$ \\
\hline
\multicolumn{11}{l}{\textit{Thinking models}} \\
Qwen3.5 Plus & 1.113 & .330/.350 & 1.480 & .133/.137 & .186 & .000/.000 & 91.7 & .843/.843 & 71.7 & .183/.183 \\
DeepSeek-V4 & 1.247 & .343/.343 & 1.677 & .017/.027 & .238 & .007/.000 & 22.0 & .150/.150 & 73.0 & .227/.220 \\
GPT-5.2 & 1.137 & .357/.390 & 1.927 & .053/.050 & .342 & .067/.067 & 53.7 & .323/.323 & 90.7 & .233/.230 \\
Gemini-3.1-Pro & 1.043 & \textbf{.487/.490} & \textbf{.410} & \textbf{.690/.690} & \textbf{.961} & \textbf{.887/.900} & \textbf{98.0} & \textbf{.920/.920} & 94.3 & \textbf{.613/.610} \\
\hline
\multicolumn{11}{l}{\textit{No-thinking models}} \\
DeepSeek-V3.2 & 1.220 & .267/.340 & 1.457 & .023/.030 & .170 & .017/.017 & 42.0 & .053/.053 & 86.3 & .240/.237 \\
Doubao-2Pro & .960 & .437/.473 & 1.150 & .267/.277 & .544 & .237/.243 & 88.0 & .760/.760 & 89.3 & .253/.253 \\
GLM-5.1 & \textbf{.940} & .353/.463 & 1.473 & .060/.077 & .193 & .020/.020 & 59.3 & .307/.307 & 93.0 & .210/.207 \\
Claude-Sonnet-4.6 & 1.020 & .417/.437 & 1.187 & .233/.260 & .647 & .237/.263 & 77.7 & .677/.673 & \textbf{96.7} & .453/.450 \\
\hline
\multicolumn{11}{l}{\textit{Layer-2 State Score over all molecular understanding model-task pairs: min/median/max = .9367/.9992/1.0000.}} \\
\hline
\end{tabular}}
\caption{Molecular understanding results. L1 reports task-specific metrics. L3 I/II reports Type-I all-pass and Type-II all-match rates. Best values per task group are bolded.}
\label{tab:molund}
\end{table*}

Checkpoint logs show that ring-count failures concentrate in ring-pattern identification (67.0\%) and total-count validation (50.4\%), while Murcko scaffold failures spike in substructure containment (73.0\%).
Thus, the benchmark localizes vague scaffold or ring-count errors to concrete graph-state commitments.

% 分子编辑部分
% 诚实地指出 Layer 1 整体偏高，对比三个子任务替换>add/delete. substitute 要求模型在 CoT 中同时处理“断键”和“成键”两个动态过程，Layer 3 的原子/质量守恒校验 (Conservation checks) 很容易在这里抓到模型“凭空变出碳原子”的幻觉
\paragraph{Molecule editing.}
Table~\ref{tab:edit_opt} summarizes molecule editing and molecular optimization in a shared task table.
Molecule editing is more constrained than open-ended generation because examples are recast from real reactant--product changes.
Consequently, Layer-1 exact match can be high, especially for add/delete operations, but this does not mean the model has preserved the full molecular state.
The structured trace tests whether the model can expose the local edit site, construct the product, and keep global invariants such as ring count and heavy-atom accounting stable.

\begin{table*}[t!]
\centering
\scriptsize
\setlength{\tabcolsep}{2.0pt}
\resizebox{\textwidth}{!}{
\begin{tabular}{l cc cc cc | cc cc cc cc}
\hline
&
\multicolumn{6}{c|}{\textbf{Molecule Editing}} &
\multicolumn{8}{c}{\textbf{Molecular Optimization}} \\
\cline{2-15}
&
\multicolumn{2}{c}{\textbf{Add}} &
\multicolumn{2}{c}{\textbf{Delete}} &
\multicolumn{2}{c|}{\textbf{Substitute}} &
\multicolumn{4}{c}{\textbf{Single}} &
\multicolumn{4}{c}{\textbf{Dual}} \\
\cline{8-15}
&
\multicolumn{2}{c}{} &
\multicolumn{2}{c}{} &
\multicolumn{2}{c|}{} &
\multicolumn{2}{c}{\textbf{Physicochemical}} &
\multicolumn{2}{c}{\textbf{Biological Target}} &
\multicolumn{2}{c}{\textbf{Physicochemical}} &
\multicolumn{2}{c}{\textbf{Biological Target}} \\
\textbf{Model} &
L1 Acc.$\uparrow$ & L3 I/II$\uparrow$
& L1 Acc.$\uparrow$ & L3 I/II$\uparrow$
& L1 Acc.$\uparrow$ & L3 I/II$\uparrow$
& L1 SR$\uparrow$ & L3$\uparrow$
& L1 SR$\uparrow$ & L3$\uparrow$
& L1 D-SR$\uparrow$ & L3$\uparrow$
& L1 D-SR$\uparrow$ & L3$\uparrow$ \\
\hline
\multicolumn{15}{l}{\textit{Thinking models}} \\
Qwen3.5 Plus & 83.3 & \textbf{.860}/.790 & 96.7 & .920/.910 & 69.0 & .813/.637 & 86.9 & .485 & 36.7 & .463 & 6.7 & .537 & 2.7 & .514 \\
DeepSeek-V4 & 92.3 & .673/.440 & 96.0 & .783/.587 & 85.3 & .780/.547 & 66.7 & .469 & 36.7 & .479 & 12.0 & .547 & 4.7 & .538 \\
GPT-5.2 & 69.3 & .203/.170 & 85.3 & .240/.237 & 60.3 & .477/.333 & 83.9 & .487 & 46.4 & .453 & 7.3 & .520 & 6.0 & .510 \\
Gemini-3.1 & \textbf{94.0} & .857/\textbf{.810} & \textbf{98.7} & \textbf{.980/.967} & \textbf{91.7} & \textbf{.950/.837} & \textbf{93.1} & \textbf{.549} & 51.4 & .530 & 12.7 & \textbf{.592} & \textbf{10.0} & \textbf{.582} \\
\hline
\multicolumn{15}{l}{\textit{No-thinking models}} \\
DeepSeek-V3.2 & 61.7 & .347/.237 & 65.0 & .227/.177 & 51.7 & .447/.287 & 76.7 & .459 & 37.5 & .443 & 7.3 & .490 & 4.7 & .487 \\
Doubao-2Pro & 81.3 & .670/.603 & 91.7 & .737/.723 & 49.3 & .627/.403 & 82.5 & .515 & 45.8 & .473 & 2.7 & .551 & 5.3 & .520 \\
GLM-5.1 & 75.3 & .617/.493 & 96.0 & .737/.710 & 54.7 & .520/.343 & 86.1 & .492 & 47.8 & .519 & 13.3 & .558 & 5.3 & .540 \\
Claude-Sonnet & 86.7 & .663/.570 & 84.7 & .630/.570 & 80.3 & .793/.660 & 91.9 & .539 & \textbf{59.4} & \textbf{.537} & \textbf{16.0} & .570 & \textbf{10.0} & .557 \\
\hline
\multicolumn{15}{l}{\textit{Layer-2 State Score over all molecule editing model-task pairs: min/median/max = .8900/.9836/1.0000.}} \\
\multicolumn{15}{l}{\textit{Layer-2 State Score over all molecular optimization paper-facing model-group pairs: min/median/max = .9520/.9922/1.0000.}} \\
\hline
\end{tabular}}
\caption{Molecule editing and molecular optimization results. Best values within each task group are bolded.}
\label{tab:edit_opt}
\end{table*}

Failures are mostly state-update errors rather than parsing errors: add/delete errors concentrate in ring-count consistency (29.8\%/27.6\%) and heavy-atom accounting (17.1\%/16.5\%), while substitution produces Type-II mismatches at product construction and final-answer fields (32.4\%/34.2\%).
This pattern is consistent with a model that can describe a plausible local transformation but loses track of the molecule-level consequences of that transformation.

% 分子优化部分
% 关注单目标vs多目标，理化优化vs靶点相关
% 点出来多目标优化的灾难：单目标时模型还有一定成功率，但到了双目标得分断崖式下跌，甚至清零
% 分析原因：化学优化是一个“牵一发而动全身”的过程。模型在 CoT 的前半段为了满足属性 A（如脂溶性）改了某个基团，结果在后半段校验属性 B 时逻辑崩溃，导致最终得到的分子虽然 Layer 1 测出来可能满足一个，但破坏了原有的 Scaffold（在 Layer 3 的 Scaffold preservation check 失败）。这证明现有的reasoning机制难以处理带有强耦合约束的化学状态机。
\paragraph{Molecular optimization.}
Molecular optimization shows a sharper constraint-coupling effect.
Models retain reasonable success on single-objective optimization, especially physicochemical properties (83.5\% avg success rate, SR), but biological target optimization is harder (45.2\% avg SR).
The collapse appears in dual-objective settings: although the marginal success rate for each objective remains about 71\%, the joint success rate drops to 9.8\% for dual physicochemical optimization and 6.1\% for dual biological-target optimization.
This indicates that models often know useful local edit heuristics, but fail when the trace must satisfy multiple coupled commitments at once.
The diagnostic value of Layer~3 is that it shows whether the generated molecule is supported by a consistent edit plan and scaffold/objective verification, not only whether one property improved.

The step logs show why this is not merely an outcome-level difficulty.
Across molecular optimization groups, functional-group change verification is almost always satisfied (about 99\% oracle-verified consistency), but scaffold-preservation verification is the weakest step, with failure rates from 73.0\% to 83.2\%.
This indicates that models can describe a local edit for one property, yet fail to keep the global molecular state stable when a second coupled constraint is imposed.
In other words, molecular optimization exposes a failure to maintain oracle-verified scaffold and objective-state consistency under coupled constraints.

% 反应预测部分
% 观察：模型很多时候在layer4 typeI(Intrinsic符号检查、配平、SMILES合法性）等方面过了，但是在typeII所代表的与真实step-level reference trace的语义对齐层面挂了。
% 分析：说明模型在反应推理时，会出现“胡扯一个化学上合法但是在特定反应体系下面不会生成的副产物或者反应路径”。这体现了模型具有一定的general chemistry syntax的能力，但是缺乏context-specific reaction mechanism的知识。

\begin{table*}[!ht]
\centering
\scriptsize
\setlength{\tabcolsep}{2.0pt}
\resizebox{\textwidth}{!}{
\begin{tabular}{l cc cc cc cc cc cc}
\hline
\textbf{Model} &
\multicolumn{2}{c}{\textbf{Product-Level}} &
\multicolumn{2}{c}{\textbf{Retrosynthesis}} &
\multicolumn{2}{c}{\textbf{Template/Mechanism}} &
\multicolumn{2}{c}{\textbf{Component Recommendation}} &
\multicolumn{2}{c}{\textbf{Condition Rank}} &
\multicolumn{2}{c}{\textbf{Yield Pred.}} \\
& L1 Acc.$\uparrow$ & L3 I/II$\uparrow$
& L1 Acc.$\uparrow$ & L3 I/II$\uparrow$
& L1 Acc.$\uparrow$ & L3 I/II$\uparrow$
& L1 Acc.$\uparrow$ & L3 I/II$\uparrow$
& L1 Acc.$\uparrow$ & L3 I/II$\uparrow$
& L1 MAE$\downarrow$ & L3 I/II$\uparrow$ \\
\hline
\multicolumn{13}{l}{\textit{Thinking models}} \\
Qwen3.5 Plus & 45.3 & .415/.253 & 18.5 & .140/.185 & 70.8 & .603/.603 & 31.8 & .313/.133 & 35.5 & .995/.110 & 0.295 & .000/.000 \\
DeepSeek-V4 & 33.3 & .252/.137 & 15.0 & .095/.120 & 76.0 & .588/.497 & 35.2 & .338/.143 & 34.0 & .985/.120 & 0.288 & .015/.000 \\
GPT-5.2 & 24.0 & .260/.160 & 8.0 & .065/.105 & 72.8 & .625/.632 & 27.8 & .278/.112 & 33.0 & \textbf{1.000}/.080 & 0.241 & .015/.000 \\
Gemini-3.1 & \textbf{52.7} & \textbf{.475/.277} & \textbf{28.5} & \textbf{.215/.280} & \textbf{86.2} & .755/.760 & \textbf{39.2} & \textbf{.388/.190} & 37.0 & .995/.125 & 0.299 & .015/.000 \\
\hline
\multicolumn{13}{l}{\textit{No-thinking models}} \\
DeepSeek-V3.2 & 18.3 & .120/.068 & 2.0 & .005/.055 & 68.0 & .515/.525 & 24.7 & .233/.100 & 34.5 & .995/\textbf{.135} & 0.303 & .010/.000 \\
Doubao-2Pro & 39.8 & .278/.182 & 12.5 & .075/.125 & 78.0 & .542/.575 & 33.7 & .330/.133 & 33.5 & .995/.120 & 0.295 & \textbf{.020}/.000 \\
GLM-5.1 & 35.7 & .307/.203 & 8.0 & .060/.110 & 81.0 & .517/.530 & 33.7 & .328/.145 & \textbf{38.0} & .990/.125 & 0.309 & .010/.000 \\
Claude-Sonnet & 42.0 & .413/.253 & 20.5 & .150/.210 & 84.0 & \textbf{.772/.777} & 35.0 & .347/.162 & 36.5 & \textbf{1.000}/.115 & \textbf{0.233} & .010/.000 \\
\hline
\multicolumn{13}{l}{\textit{Layer-2 State Score over all reaction prediction paper-facing model-group pairs: min/median/max = .7734/.9967/1.0000.}} \\
\hline
\end{tabular}}
\vspace{-0.2cm}
\caption{Reaction prediction results. Best non-degenerate values within each task group are bolded.}
\label{tab:rxnpred}
\end{table*}

\paragraph{Reaction prediction.}
Reaction prediction separates surface chemical syntax from context-specific reaction commitments.
Figure~\ref{fig:diagnostic_case} shows one product-level case where the product is correct but the \texttt{RXN\_TYPE} commitment disagrees with the benchmark-defined state.
The clearest aggregate case is condition ranking: Type-I validity reaches 99.4\%, but Type-II benchmark-state agreement is only 11.6\%.
Component recommendation shows a similar drop from intrinsic symbolic validity to benchmark-state agreement, suggesting that many failures are chemically plausible but benchmark-inconsistent commitments under the provided reaction context.
In other words, models can often produce syntactically valid products, rankings, or component lists, but fail to bind those outputs to the specific reaction context.
This is exactly where outcome-only evaluation is least informative: a product or option may look chemically reasonable, while the structured trace reveals that the model selected the wrong reaction abstraction, condition order, or component rationale.

\subsection{Cross-Task Insights}

\paragraph{Template following and outcome accuracy do not imply valid structured reasoning.}
Models adopt the requested formal style far more reliably than they maintain structured chemical commitments: average Layer-2 State Scores are $\geq 0.970$ across task families, but Layer-3 scores drop sharply.
Molecular understanding averages only 0.310/0.319 in Type-I/Type-II Layer~3, reaction prediction averages 0.386/0.226, and even molecule editing drops from 0.970 in Layer~2 to 0.648/0.543 in Layer~3.
The same separation appears when final answers are correct: SMILES equivalence reaches 86.9\% Layer-1 accuracy but only 29.9\% Layer-3 Type-II all-match, showing that structured rationales can be grammatical and outcome-correct while still failing intermediate commitments.

\paragraph{Generic chemical heuristics fail under grounding and composition.}
Step-level verification shows that many failures reflect poor grounding rather than a complete absence of chemical knowledge.
Condition ranking traces can be formally complete and 99.4\% Type-I valid while showing only 11.6\% benchmark-state agreement; molecular optimization similarly succeeds on single objectives but falls to 9.8\%/6.1\% joint success in dual-objective settings.
Molecule editing shows the same pattern when substitution must coordinate bond breaking, fragment insertion, product construction, and global consistency in one state transition.
Thus, current LLMs use reusable chemical associations, but struggle when they must compose and ground them in the exact atoms, scaffolds, conditions, and constraints of the instance.

\paragraph{Maintaining structured chemical commitments is the bottleneck.}
Across task families, performance drops when models must update persistent commitments rather than recognize static patterns, as in ring counting, scaffold extraction, molecular-optimization scaffold consistency, and condition ranking.
Together, these findings suggest that current LLMs remain stronger at local chemical description than at continuous molecular or reaction-state tracking.

\begin{table}[!t]
\centering
\footnotesize
\setlength{\tabcolsep}{2.4pt}
\resizebox{\columnwidth}{!}{
\begin{tabular}{lccc}
\hline
\textbf{Group} & \textbf{Direct} & \textbf{Template} & \textbf{+Anchor} \\
\hline
\multicolumn{4}{l}{\textit{Molecular understanding}} \\
Functional-group MAE$\downarrow$ & 1.380 & 1.220 & \textbf{1.120} \\
Ring-count MAE$\downarrow$ & \textbf{1.340} & 1.457 & 1.450 \\
Murcko scaffold Tanimoto$\uparrow$ & 0.157 & 0.170 & \textbf{0.184} \\
Ring-system scaffold accuracy$\uparrow$ & 0.253 & 0.420 & \textbf{1.000} \\
SMILES equivalence accuracy$\uparrow$ & 0.637 & 0.863 & \textbf{1.000} \\
\hline
\multicolumn{4}{l}{\textit{Molecular optimization}} \\
Physicochemical single-objective SR$\uparrow$ & 36.1 & 76.7 & \textbf{81.1} \\
Biological-target single-objective SR$\uparrow$ & 26.9 & 37.5 & \textbf{37.8} \\
Physicochemical dual-objective D-SR$\uparrow$ & 0.0 & \textbf{7.3} & 5.3 \\
Biological-target dual-objective D-SR$\uparrow$ & 0.7 & 4.7 & \textbf{6.7} \\
Avg. success rate$\uparrow$ & 15.9 & 31.5 & \textbf{32.7} \\
\hline
\end{tabular}}
\caption{Prompt ablation on DeepSeek-V3.2. Molecular understanding is reported with task-specific metrics (MAE, Tanimoto, accuracy). For optimization, SR/D-SR follow Table~\ref{tab:edit_opt} and denote single-/dual-objective success rates (\%). +Anchor injects safe intermediate states rather than final answers.}
\vspace{-0.1cm}
\label{tab:prompt_ablation}
\end{table}

\subsection{The Role of Reasoning Scaffolds}
\label{sec:reasoning_scaffolds}

We probe whether the failures above reflect missing chemical knowledge or difficulty planning and maintaining a reliable reasoning path using a prompt ablation on DeepSeek-V3.2.
The ablation compares \textit{Direct}, \textit{Template}, and \textit{Template+Anchor}, where anchors add safe intermediate states without revealing the answer.
The template alone substantially improves molecular optimization (avg success rate 15.93$\rightarrow$31.54) and many molecular-understanding tasks, suggesting that the model benefits from explicit intermediate commitments.

Safe intermediate anchors further improve molecular understanding, especially ring-system scaffold and SMILES equivalence, and raise molecular-optimization performance to 32.72.
Overall, the ablation suggests that explicit reasoning scaffolds help models expose and preserve chemically meaningful intermediate commitments, but the main bottleneck remains maintaining these commitments over long structured traces---the capability targeted by Layer~3.

%% file: Contents/5_Conclusion.tex
\section{Conclusion and Future Work}

We introduced \textsc{ChemCoTBench-V2}, a rule-verifiable benchmark for diagnosing structured chemical reasoning through verifier-addressable intermediate commitments across 5,620 samples and 18 reporting tasks.
By distilling natural CoT patterns into expert-refined templates and combining them with deterministic chemistry verifiers and verified benchmark-state traces, \ours{} separates final-answer correctness, template adherence, and step-wise verifier correctness.
Experiments show that frontier LLMs often produce well-formatted reasoning traces while failing chemically meaningful intermediate checks, especially when tasks require persistent molecular or reaction-state commitments.
These results suggest that current models remain stronger at local chemical description than at reliable multi-step chemical state updates.
Future work should extend the reference construction pipeline to broader reaction regimes and 3D molecular settings, and use the localized failure signals to guide training, prompting, and tool-augmented reasoning systems.

\clearpage
\section*{Limitations}

\ours{} is designed for rule-verifiable process-level reasoning on 2D molecular and reaction representations. Extending the same diagnostic protocol to settings such as 3D conformational reasoning, quantum chemistry, laboratory procedure planning, protein--ligand interaction modeling, and long-horizon synthesis is a natural direction for future work, but would require additional task-specific state definitions and verification criteria. Within the current scope, expert-designed templates and rule-based checks provide stable and reproducible evaluation; Type-II agreement should be interpreted as benchmark-state agreement for closed-answer tasks, while open-ended settings require oracle-verifiable constraints or task-specific state definitions rather than exhaustive rationale judgments.

Our evaluation should not be interpreted as unrestricted sentence-level judging of free-form CoT.
Instead, \ours{} evaluates structured reasoning commitments distilled from natural CoT traces and refined by chemistry experts.
Candidate reference traces are constructed with access to the final answer, so they should be viewed as verified benchmark-state trajectories rather than unique human reasoning processes.
We mitigate post-hoc rationalization risk by fixing templates before reference construction, applying deterministic rule checks, resolving multi-model conflicts, and auditing a stratified expert sample.
Future work should expand task-wise expert validation and support multiple accepted state trajectories for semantically softer tasks.

% \section*{Acknowledgments}

% We thank the members of our research group for discussions on task design, data validation, and evaluation analysis.
% We also acknowledge the open-source chemistry ecosystem and public molecular and reaction resources that made this benchmark possible, including RDKit and the public datasets used in our construction pipeline.

%% file: Contents/appendix.tex
\appendix

\section{Dataset Construction Details}
\label{sec:appendix}

\newcommand{\appwidefig}[4]{%
\begin{figure*}[htbp]
\centering
\includegraphics[width=#1\textwidth]{#2}
\caption{#3}
\label{#4}
\end{figure*}
}

\newcommand{\appcolfig}[4]{%
\begin{figure}[htbp]
\centering
\includegraphics[width=#1\columnwidth]{#2}
\caption{#3}
\label{#4}
\end{figure}
}

\newcommand{\appresultpair}[5]{%
\begin{figure*}[htbp]
\centering
\begin{minipage}[t]{0.48\textwidth}
\centering
\includegraphics[width=\linewidth]{#2}
\vspace{2pt}
\textbf{Layer 1}
\end{minipage}
\hfill
\begin{minipage}[t]{0.48\textwidth}
\centering
\includegraphics[width=\linewidth]{#3}
\vspace{2pt}
\textbf{Layer 3}
\end{minipage}
\caption{#4}
\label{#5}
\end{figure*}
}
% ------------------------------------------

\subsection{Data Sources and Filtering}
\label{sec:app_data_sources}

The benchmark is constructed from public molecular resources, reaction corpora, matched molecular-pair collections, and task-specific reaction-condition datasets.
Molecular-understanding instances are sampled from public compound collections such as PubChem~\cite{kim2016pubchem}, ChEMBL~\cite{gaulton2017chembl}, and ZINC~\cite{irwin2012zinc}, followed by RDKit-based standardization and task-specific label generation.
Molecule-editing instances are derived from atom-mapped organic reactions in Schneider 50K, a USPTO-derived reaction corpus, by extracting the structural change from the main reactant to the main product.
Molecular-optimization instances are built from matched molecular pairs and property oracles for physicochemical and bioactivity objectives.
Reaction-prediction instances are assembled from reaction-product, retrosynthesis, reaction-template, mechanism, reaction-component, condition-ranking, and yield-prediction data pools.

All candidate instances are normalized with RDKit where applicable.
We remove invalid SMILES, ambiguous reaction records, answer-leaking metadata, and samples whose labels cannot be made consistent with the task definition.
For tasks requiring exact molecular comparison, canonical SMILES or main-fragment canonicalization is used before labels or metrics are computed.

\subsection{Reporting Tasks and Fine-Grained Tasks}
\label{sec:app_reporting_tasks}

The main paper reports results at 18 task groups to keep the experimental tables readable.
These groups cover 31 active fine-grained chemical tasks.
This grouping only affects presentation: all models are evaluated on the same 5,620 active samples, and the reported scores are sample-weighted aggregations over the corresponding fine-grained tasks.
Within each task family, the active benchmark is intentionally balanced at the finest evaluation granularity rather than dominated by a few large subtasks.

% table position changed to [htbp]
\begin{table*}[htbp]
\centering
\small
\setlength{\tabcolsep}{5pt}
\begin{tabular}{p{0.23\textwidth} p{0.45\textwidth} p{0.16\textwidth} r}
\hline
\textbf{Family} & \textbf{Reporting tasks} & \textbf{Fine-grained sampling} & \textbf{Total} \\
\hline
Molecule editing &
Add, Delete, Substitute &
3 tasks $\times$ 300 &
900 \\
Molecular understanding &
Functional Group, Ring Count, Murcko Scaffold, Ring-System Scaffold, SMILES Equivalence &
5 tasks $\times$ 300 &
1,500 \\
Reaction prediction &
Product-Level Prediction, Retrosynthesis, Template/Mechanism Reasoning, Component Recommendation, Condition Ranking, Yield Prediction &
11 tasks $\times$ 200 &
2,200 \\
Molecular optimization &
PhysChem-Single, BioTarget-Single, PhysChem-Dual, BioTarget-Dual &
6 single tasks $\times$ 120; 6 dual tasks $\times$ 50 &
1,020 \\
\hline
\textbf{Total} & \textbf{18 reporting tasks} & \textbf{31 fine-grained tasks} & \textbf{5,620} \\
\hline
\end{tabular}
\caption{Active benchmark composition. Reporting tasks are used for paper-facing tables, while all evaluation is run over the underlying fine-grained tasks.}
\label{tab:app_task_composition}
\end{table*}

\appwidefig{0.9}{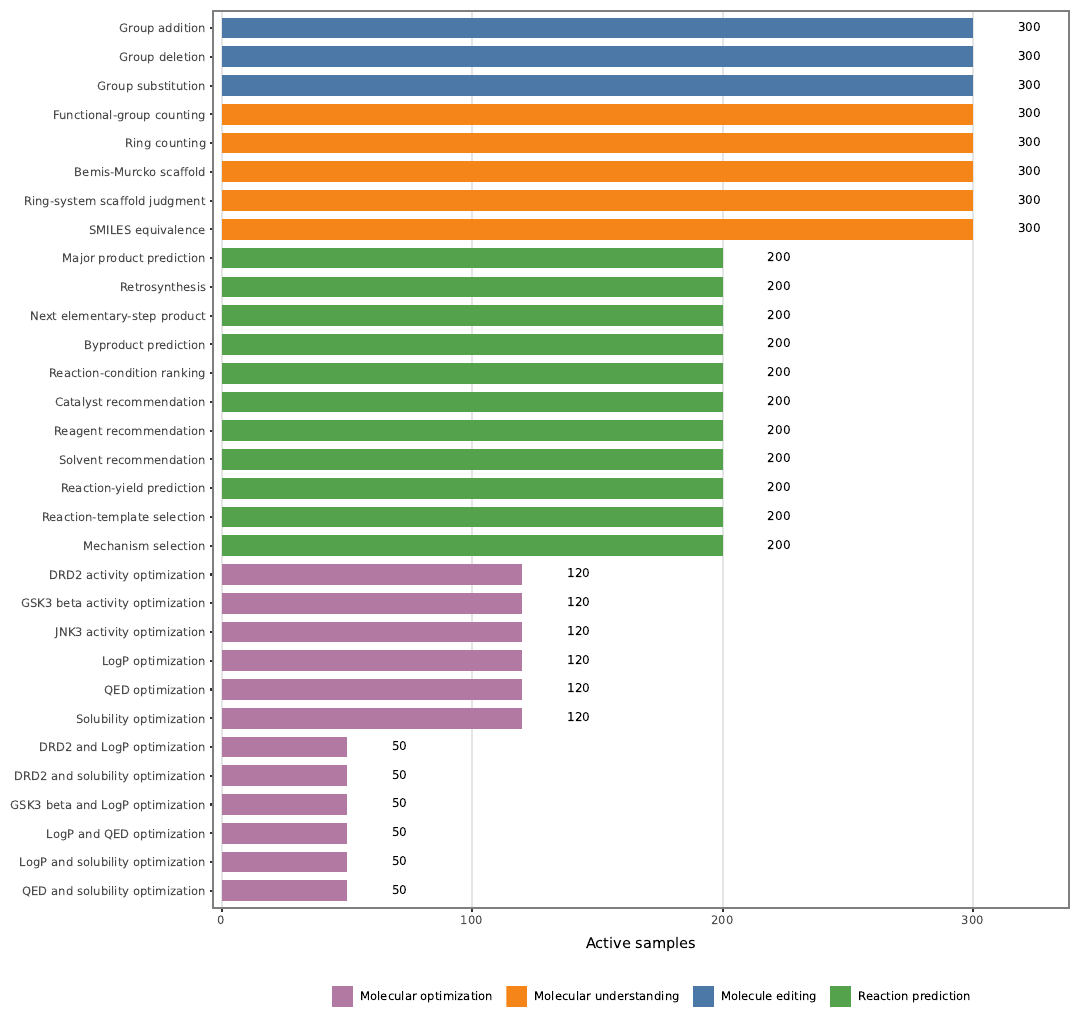}
{Active sample counts for the 31 fine-grained chemical tasks. The distribution is uniform within each task family: 300 samples per molecule-editing and molecular-understanding task, 200 samples per reaction-prediction task, 120 samples per single-objective molecular-optimization task, and 50 samples per dual-objective molecular-optimization task.}
{fig:app_fine_task_distribution}

\paragraph{Molecule editing.}
The three reported groups are the addition of a reaction-derived group, the deletion of a protecting group or substituent, and the substitution of a leaving group with a new group.
Each contains 300 samples and is scored by exact molecular matching.

\paragraph{Molecular understanding.}
The five reported groups are functional-group counting, ring counting, Bemis--Murcko scaffold extraction, ring-system scaffold judgment, and SMILES equivalence.
The first two are count tasks scored by mean absolute error.
The scaffold extraction task is scored by molecular similarity, while ring-system scaffold judgment and SMILES equivalence are scored by exact accuracy.
The SMILES-equivalence group contains two internal variants: equivalent SMILES permutations and chemically perturbed non-equivalent SMILES.

\paragraph{Reaction prediction.}
The six reported groups are product-level prediction, retrosynthesis, template and mechanism reasoning, reaction-component recommendation, condition ranking, and numerical yield prediction.
Product-level prediction combines major-product prediction, byproduct prediction, and next elementary-step product prediction.
Reaction-component recommendation combines catalyst, reagent, and solvent recommendations.
Yield prediction is scored by mean absolute error; the other reaction-prediction groups are scored by top-1 accuracy.

\paragraph{Molecular optimization.}
The four reported groups separate single-objective and dual-objective optimization, and also separate physicochemical objectives from bioactivity objectives.
The physicochemical objectives include LogP, QED, and solubility.
The bioactivity objectives include DRD2, JNK3, and GSK3$\beta$ activity.
Single-objective tasks are scored by success rate, while dual-objective tasks are scored by dual success rate.

\subsection{From the Construction Pool to the Active Benchmark}
\label{sec:app_reduce}

The initial construction pool contained 12,600 samples across the fine-grained tasks.
We selected a compact active benchmark of 5,620 samples to make full-model evaluation computationally feasible while preserving balanced coverage across task families and fine-grained tasks.

\appcolfig{0.95}
{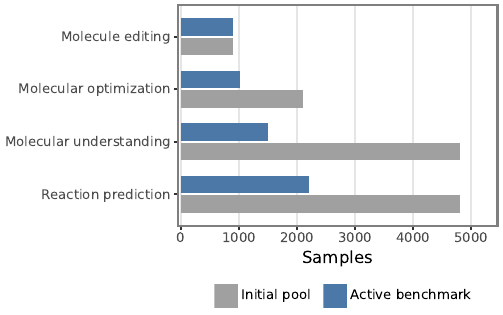}
{Reduction from the initial construction pool to the active evaluation benchmark. The final active benchmark contains 5,620 samples selected from a 12,600-sample construction pool.}
{fig:app_reduction}

Molecule editing contains 300 samples for each edit type, molecular understanding contains 300 samples for each active group, reaction prediction contains 200 samples for each fine-grained task, and molecular optimization contains 120 samples for each single-objective task and 50 samples for each dual-objective task.

\subsection{Molecule Editing from Reaction-Derived Structural Changes}
\label{sec:app_moledit}

Traditional molecule-editing tasks often ask a model to modify a molecule from a natural-language instruction, but the instruction may be weakly grounded in a real chemical transformation.
In contrast, our molecule-editing instances are derived from real organic reactions.
Starting from atom-mapped reactions in Schneider 50K, we identify the main reactant, extract the largest product fragment, canonicalize both molecules, and treat the source-to-target difference as a localized molecular edit.
This produces three chemically interpretable edit types: adding a group, deleting a group, and substituting one group for another.

\paragraph{Filtering and edit extraction.}
Candidate source-target pairs are filtered with RDKit before instruction generation.
We retain pairs in which both molecules are valid, the product is related but not nearly identical to the source, the heavy-atom difference is bounded, and the source molecule is not overly trivial.
In practice, we use a Tanimoto-similarity window of approximately $[0.35,0.95]$, a heavy-atom-difference range of $[1,15]$, and a minimum source-molecule complexity threshold of 30.
These filters remove invalid reactions, trivial perturbations, and large molecular reorganizations.

\paragraph{Edit-type classification.}
The edit type is determined from the reaction-derived structural change rather than from net heavy-atom change alone.
This is necessary because many substitutions increase the molecule size.
For example, in a Suzuki coupling, an aryl bromide can be replaced by a larger aryl group; the product gains heavy atoms, but the chemical operation is still substitution because a leaving group is replaced.
We therefore distinguish: (i) addition, where a new group is introduced without an explicit leaving group; (ii) deletion, where a protecting group or substituent is removed, commonly through deprotection or hydrolysis; and (iii) substitution, where a leaving group disappears, and a new group enters.

\paragraph{Site-specific instruction generation.}
The extracted reaction pairs do not contain natural-language edit instructions.
We use GPT-5.4 to produce concise site-specific instructions from the source molecule, target molecule, reaction class, and structural-change metadata.
Each candidate is sampled multiple times.
We retain only candidates with successful parsing, high confidence, and high instruction agreement, measured by pairwise word-overlap similarity between independently generated instructions.
Instructions that describe reagents or reaction conditions rather than the molecular edit itself are filtered out.

\paragraph{Balancing and deduplication.}
After quality filtering, samples are deduplicated by source-target pair.
For addition and substitution, repeated instruction templates are capped to preserve diversity.
For deletion, repeated deprotection instructions are chemically unavoidable, so we use progressive filling: first, take the best instance for each unique instruction, then the second-best instance, and so on until 300 samples are reached.
Instruction agreement is used only as a quality filter.

\appwidefig{0.9}
{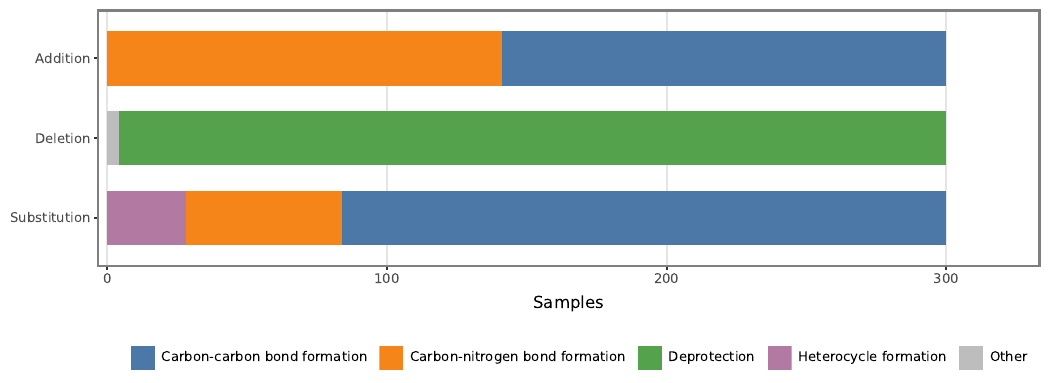}
{Reaction-class composition of the molecule-editing active set. Each edit type contains 300 samples. Average source-target Tanimoto similarities are 0.908 for addition, 0.916 for deletion, and 0.846 for substitution.}
{fig:app_moledit_stats}

The resulting editing tasks are therefore not string-editing puzzles.
Each instance specifies a reaction-derived local graph update, while the formal reasoning trace later checks anchor identification, fragment identification, product construction, heavy-atom accounting, and ring-count consistency.

\subsection{Condition Ranking Construction and Label Shuffling}
\label{sec:app_condition_ranking}

The reaction-condition ranking task tests whether a model can compare experimental conditions rather than merely classify a reaction.
Each instance contains one reaction and three candidate condition sets.
The target output is a ranking of the three conditions by expected yield.
The formal template decomposes the reasoning into six steps: reaction-class identification, decision-factor selection, three pairwise condition comparisons, pairwise preference construction, global ranking, and top-2 support.

During reference-trace construction, the three conditions are presented in ground-truth yield order so that the reference trace can be verified and repaired reliably.
Consequently, the unshuffled reference data always has condition 1 as the best-yield condition, condition 2 as the middle-yield condition, and condition 3 as the worst-yield condition.
This ordering is useful for constructing reference traces, but it would create a shortcut for evaluation: a model could always output the order 1--2--3 without reading the condition content.

To remove this label-order bias, the active evaluation set randomly permutes the three condition labels and recomputes the ground-truth ranking under the new labels.
In the current 200-sample active set, the six possible rankings are approximately balanced:
2--1--3: 39,
1--2--3: 37,
1--3--2: 32,
3--2--1: 31,
2--3--1: 31, and
3--1--2: 30.

\subsection{SMILES Equivalence Merge}
\label{sec:app_smiles_equiv}

The molecular-understanding suite contains two SMILES-equivalence variants.
The first asks whether two different SMILES strings represent the same molecule after canonicalization.
The second asks whether a chemically perturbed SMILES represents a different molecule.
For reporting, these are merged into one SMILES Equivalence task because both variants test whether a model can decide if two molecular strings denote the same chemical structure.

The active 300-sample set is sampled to keep the two variants approximately balanced, resulting in 157 equivalent-SMILES examples and 143 chemically perturbed examples.

\subsection{Expert Validation of the Verifier}
\label{app:expert_validation}
To validate the reference construction protocol, we randomly sampled 300 step-level traces across the four task families and asked 3 expert chemists with experience in organic chemistry and cheminformatics to independently judge each step as correct, incorrect, or ambiguous. Disagreements were adjudicated by majority vote. The deterministic verifier agreed with the adjudicated expert label on 87.4\% of step judgments (Cohen's $\kappa$ = 0.74), and human-human agreement was 90.1\% (Cohen's $\kappa$ = 0.79). Disagreement cases were concentrated in tasks with intrinsic path multiplicity, especially molecular optimization, condition ranking, and retrosynthesis.

\section{Fine-Grained Evaluation Results}
\label{sec:appendix_fine_results}

This section reports fine-grained results for the 31 active implementation subtasks.
Layer~2 is omitted because it measures template-state compliance rather than final chemical correctness or step-wise reasoning quality; its scores are already summarized in the main experiments.
For each entry below, L1 is the native outcome metric for that subtask, and L3 is the process metric.
For molecule editing, molecular understanding, and reaction prediction, L3 is reported as Type-I all-pass / Type-II all-match.
For molecular optimization, L3 is the average oracle-verified optimization state score.
The tables are intended to replace the earlier color-normalized heatmaps with directly readable values.

\subsection{Layer-3 Verifier Checkpoints by Subtask}
\label{sec:app_l3_checkpoints}

This subsection lists the concrete Layer-3 checkpoints used by the released verifier implementations.
For molecule editing, molecular understanding, and reaction prediction, Type-I checks validate the reasoning trace internally, while Type-II checks compare parsed state fields with the verified benchmark-state trace.
For molecular optimization, the verifier instead averages five oracle-checkable state claims over the generated molecule and does not use Type-II path matching.
Tables~\ref{tab:app_l3_moledit}--\ref{tab:app_l3_molopt} are placed inline with the text so that the reader can inspect each task family together with its explanation.
The group repeated checks rather than repeating identical logic for every row.
Implementation flag names are included only as traceability anchors; the main table cells describe the actual verifier logic.

\paragraph{Molecule editing.}
Table~\ref{tab:app_l3_moledit} shows the local-edit checks used for Add, Delete, and Substitute.
All three subtasks verify an edit anchor, the edited fragments, the generated product, and global heavy-atom/ring accounting; the task-specific difference is whether the edit introduces, removes, or swaps fragments.

\begin{table*}[!ht]
\centering
\small
\setlength{\tabcolsep}{3pt}
\renewcommand{\arraystretch}{1.08}
\begin{tabular}{p{0.13\textwidth} p{0.58\textwidth} p{0.21\textwidth}}
\hline
\textbf{Subtask} & \textbf{Type-I checkpoint logic} & \textbf{Type-II matched states} \\
\hline
Add &
Anchor grounding: the 1-based anchor index must be within the RDKit atom range and the atom element must match the declared element (\texttt{s1\_idx\_valid}, \texttt{s1\_element\_match}). Fragment validity: the added fragment SMILES must be RDKit-parseable and its declared heavy-atom count must equal RDKit's count (\texttt{s2\_frag\_valid}, \texttt{s2\_heavy\_atoms\_ok}). Product/count consistency: the product SMILES must parse; source/product heavy-atom counts and ring counts must equal RDKit values; both declared deltas must equal product minus source (\texttt{s3\_product\_valid}, \texttt{s4\_*}, \texttt{s5\_*}). &
Anchor, leaving marker, added fragment, product SMILES, heavy-atom fields, ring fields, and final answer. \\
\hline
Delete &
Shared anchor, product-validity, heavy-atom arithmetic, and ring-arithmetic checks are the same as Add. The task-specific fragment check requires the removed-group SMILES to be RDKit-parseable and its declared heavy-atom count to equal RDKit's group count (\texttt{s2\_group\_valid}, \texttt{s2\_heavy\_atoms\_ok}). &
Anchor, removed group, product SMILES, heavy-atom fields, ring fields, and final answer. \\
\hline
Substitute &
Anchor grounding is the same as Add/Delete. The verifier then checks both edit sides: removed-group and added-fragment SMILES must each parse, and each declared heavy-atom count must match RDKit (\texttt{s2\_remove\_*}, \texttt{s3\_add\_*}). The product must parse, and source/product heavy-atom counts, ring counts, and both deltas must match RDKit and arithmetic (\texttt{s4\_product\_valid}, \texttt{s5\_*}, \texttt{s6\_*}). &
Anchor, removed group, added fragment, product SMILES, heavy-atom fields, ring fields, and final answer. \\
\hline
\end{tabular}
\captionof{table}{Layer-3 checkpoints for molecule-editing subtasks. Asterisks denote the grouped implementation flags for the corresponding count/delta checks.}
\label{tab:app_l3_moledit}
\end{table*}

\paragraph{Molecular understanding.}
Table~\ref{tab:app_l3_molunderstand} covers the five structure-understanding subtasks.
These checks mainly ask whether symbolic claims in the trace agree with RDKit canonicalization, SMARTS matching, ring/scaffold computation, and the benchmark target state.

\begin{table*}[!ht]
\small
\setlength{\tabcolsep}{3pt}
\renewcommand{\arraystretch}{1.08}
\begin{tabular}{p{0.15\textwidth} p{0.56\textwidth} p{0.20\textwidth}}
\hline
\textbf{Subtask} & \textbf{Type-I checkpoint logic} & \textbf{Type-II matched states} \\
\hline
Functional-group detection &
The declared SMARTS must be RDKit-parseable and semantically aligned with the target group. The reported SMARTS match count must equal RDKit substructure matching; the accepted count must equal that match count; the final answer must equal the accepted count (\texttt{S1\_syntax}, \texttt{S1\_semantic}, \texttt{S2\_count}, \texttt{S3\_arith}, \texttt{S4\_ans}). &
SMARTS/count state and final count. \\
\hline
Ring counting &
The target-ring SMARTS must parse and match the requested ring type. The declared total SSSR ring count must equal RDKit \texttt{CalcNumRings}; enumerated SMARTS matches must equal RDKit matches; accepted count must equal enumerated matches; rejected count must equal total rings minus accepted matches (\texttt{S1\_*}--\texttt{S5\_*}). &
SMARTS, total rings, match count, accepted/rejected counts, and final answer. \\
\hline
Murcko scaffold extraction &
The source ring count must equal RDKit. The declared scaffold SMILES must parse, match RDKit's Bemis--Murcko scaffold, have the correct RDKit ring count, preserve the declared source/scaffold ring relation, and be a source substructure when applicable (\texttt{S1\_mol\_rings}, \texttt{S2\_*}, \texttt{S3\_scaf\_rings}, \texttt{S4\_ring\_match}, \texttt{S5\_substructure}). &
Scaffold SMILES, ring counts, ring-match decision, substructure decision, and final scaffold answer. \\
\hline
Ring-system scaffold judgment &
The molecule ring count and candidate-scaffold ring count must match RDKit. The non-ring-linker claim must agree with RDKit graph comparison, and the final yes/no answer must follow from the verified ring-system relation (\texttt{s1\_mol\_rings}, \texttt{s2\_scaf\_rings}, \texttt{s3\_non\_ring}, \texttt{s4\_predict}). &
Ring counts, non-ring relation, and final ring-system-scaffold judgment. \\
\hline
SMILES equivalence &
Both input SMILES must parse and canonicalize. For each molecule, declared canonical SMILES and molecular formula must equal RDKit outputs. The formula-match claim must equal formula equality; the identity claim must equal canonical graph equality; the final Same/Different prediction must follow from that identity claim (\texttt{S1\_*}--\texttt{S5\_*}). &
Canonical states, formula states, identity decision, and final answer. \\
\hline
\end{tabular}
\captionof{table}{Layer-3 checkpoints for molecular-understanding subtasks.}
\label{tab:app_l3_molunderstand}
\end{table*}

\paragraph{Reaction prediction: products, retrosynthesis, templates, and mechanisms.}
Table~\ref{tab:app_l3_rxn_core} lists the reaction subtasks where the model must predict a product/reactant state or choose a mechanistic abstraction.
The Type-I side checks grounded functional groups, reaction classes, mechanisms, parseability, charge/atom conservation, bond changes, and template consistency; Type-II then compares the closed state fields with the verified trace.

\begin{table*}
\small
\setlength{\tabcolsep}{3pt}
\renewcommand{\arraystretch}{1.08}
\begin{tabular}{p{0.15\textwidth} p{0.56\textwidth} p{0.20\textwidth}}
\hline
\textbf{Subtask} & \textbf{Type-I checkpoint logic} & \textbf{Type-II matched states} \\
\hline
Forward product prediction &
Functional-group evidence must be grounded in the reactants. The coarse reaction type must be allowed and exact; the mechanism keyword must be compatible with that type. The predicted product SMILES must parse, and the declared bond-formation state must agree with the reactant/product graph change (\texttt{S1\_fg\_grounding}, \texttt{S2\_rxn\_type\_exact}, \texttt{S3\_mechanism}, \texttt{S4\_mol\_valid}, \texttt{S5\_bond\_formed}). &
Reaction type, product, and answer. \\
\hline
Byproduct prediction &
Functional-group evidence and reaction type must be grounded. The atomic-delta statement must be compatible with the reaction; the leaving-fragment/byproduct precursor must parse; the fragment must appear in the reactant context; fragment elements must be coherent with the predicted byproduct (\texttt{S1\_*}--\texttt{S6\_*}). &
Reaction type, byproduct SMILES, and answer. \\
\hline
Next elementary-step product prediction &
Reactant charge state must parse and be correct; the elementary-mechanism label must be valid; the bond-change statement must match the elementary step; product SMILES must parse; product charge must match RDKit; charge balance and atom conservation must hold between elementary-step reactants and product (\texttt{S1\_*}--\texttt{S7\_*}). &
Elementary-step product and final answer. \\
\hline
Retrosynthesis &
Functional-group evidence must be grounded in the product. Reaction type and mechanism must be consistent. All predicted reactant SMILES must parse; the bond-breaking claim must match the product-to-reactant graph change; a forward self-check must plausibly reconstruct the product from the proposed reactants (\texttt{S1\_*}--\texttt{S6\_*}). &
Reaction type, reactants, and answer. \\
\hline
Reaction-template selection &
Declared bond changes must be valid for the reactant/product pair. Reaction type and mechanism must be consistent. The proposed reaction SMARTS must parse, match the benchmark template abstraction when required, and support the selected option (\texttt{S1\_bond\_changes\_valid}, \texttt{S2\_*}--\texttt{S5\_*}). &
Reaction type, selected option, and answer. \\
\hline
Mechanism selection &
Candidate mechanism SMARTS must parse. The reaction type must be exact; the elimination logic must remove incompatible candidates; the remaining-candidate count must equal the candidate set minus eliminated options; the selected mechanism must be one of the remaining valid options (\texttt{S1\_smarts\_parseable}, \texttt{S2\_*}--\texttt{S5\_*}). &
Reaction type, mechanism option, and answer. \\
\hline
\end{tabular}

\captionof{table}{Layer-3 checkpoints for product, retrosynthesis, template, and mechanism reaction subtasks.}
\label{tab:app_l3_rxn_core}
\end{table*}

\paragraph{Reaction prediction: components, conditions, and yield.}
Table~\ref{tab:app_l3_rxn_recommend} covers the remaining reaction subtasks.
Here, the verifier checks whether the model's recommendation or ranking is grounded in the reaction class and declared decision factors, then compares the answer-level state with the benchmark label.

\begin{table*}
\small
\setlength{\tabcolsep}{3pt}
\renewcommand{\arraystretch}{1.08}
\begin{tabular}{p{0.15\textwidth} p{0.56\textwidth} p{0.20\textwidth}}
\hline
\textbf{Subtask} & \textbf{Type-I checkpoint logic} & \textbf{Type-II matched states} \\
\hline
Catalyst recommendation &
Reaction class must be exact; the core-transformation statement must be consistent with the reaction; catalyst role must be explicitly identified; catalyst class must be grounded in that role; predicted catalyst must be self-consistent with the class, RDKit-parseable, and identical to the answer field (\texttt{S1\_*}--\texttt{S7\_*}). &
Reaction class, catalyst SMILES, and answer. \\
\hline
Reagent recommendation &
Reaction class and core transformation must be grounded. The reagent slot, component mode, and reagent class must be valid; predicted reagent must be self-consistent with component mode and class, not merely copied from reactants, charge-balanced when ionic, RDKit-parseable, and identical to the answer field (\texttt{S1\_*}--\texttt{S12\_*}). &
Reaction class, reagent SMILES, and answer. \\
\hline
Solvent recommendation &
Reaction class and core transformation must be grounded. Declared proticity and polarity requirements must be consistent with the solvent strategy; predicted solvent SMILES must parse; the answer field must match the predicted solvent (\texttt{S1\_*}, \texttt{S2\_*}, \texttt{S4\_*}--\texttt{S7\_*}). &
Reaction class, solvent SMILES, and answer. \\
\hline
Condition ranking &
The decision factor must be an allowed condition field. Pair-difference records must cover pairs 1/2, 1/3, and 2/3; pairwise preferences must cover all unordered pairs exactly once; preferences must form an acyclic total order; the ranking must be a permutation of the three condition labels; top-2 support must agree with the ranking and decision factor (\texttt{S1\_*}--\texttt{S6\_*}). &
Reaction class, shuffled-condition ranking, and answer. \\
\hline
Yield prediction &
Reaction class, halide identity, nucleophile type/form, and ligand-system class must use valid labels. Predicted yield must be numeric, within 0--100, and identical to the answer field (\texttt{S1\_rxn\_class}, \texttt{S2\_halide\_type}, \texttt{S3\_nucleophile\_fmt}, \texttt{S4\_ligand\_class\_fmt}, \texttt{S5\_yield\_numeric}, \texttt{S6\_yield\_range}, \texttt{S7\_answer\_consistent}). &
Reaction class, yield value, and answer, using the task's numeric tolerance. \\
\hline
\end{tabular}
\captionof{table}{Layer-3 checkpoints for reaction component recommendation, condition ranking, and yield prediction.}
\label{tab:app_l3_rxn_recommend}
\end{table*}

\paragraph{Molecular optimization.}
Table~\ref{tab:app_l3_molopt} is intentionally shared across all 12 molecular-optimization subtasks.
The objective oracle used for Layer~1 changes by task, but Layer~3 always checks the same five exposed optimization states: scaffold extraction, edit-plan validity, product validity, scaffold preservation, and functional-group change consistency.

\begin{table*}
\small
\setlength{\tabcolsep}{3pt}
\renewcommand{\arraystretch}{1.08}
\begin{tabular}{p{0.24\textwidth} p{0.17\textwidth} p{0.50\textwidth}}
\hline
\textbf{Shared subtasks} & \textbf{Checkpoint} & \textbf{Oracle logic} \\
\hline
\texttt{logp}, \texttt{qed}, \texttt{solubility}, \texttt{drd}, \texttt{jnk}, \texttt{gsk}, \texttt{logp\_qed}, \texttt{logp\_solubility}, \texttt{qed\_solubility}, \texttt{drd\_logp}, \texttt{drd\_solubility}, \texttt{gsk\_logp} &
Scaffold extraction &
The claimed scaffold must match the RDKit Bemis--Murcko scaffold of the source molecule after the task's scaffold-normalization rules (\texttt{S1\_scaffold}). \\
\hline
Same 12 subtasks &
Edit-plan validity &
The declared removed and added functional-group fragments must be valid fragment SMILES or \texttt{none}, and at least one side of the edit must be non-\texttt{none} (\texttt{S2\_edit\_plan}). \\
\hline
Same 12 subtasks &
Product validity &
The predicted optimized molecule must be RDKit-parseable as a complete molecule (\texttt{S3\_product}). \\
\hline
Same 12 subtasks &
Scaffold preservation &
The trace's yes/no scaffold-preservation claim must equal the oracle comparison between the source and predicted Murcko scaffolds (\texttt{S4\_scaffold}). \\
\hline
Same 12 subtasks &
Functional-group change &
The claimed removed/added fragments must be reflected in the source-to-product structural difference detected by the verifier (\texttt{S5\_fg}). \\
\hline
\end{tabular}
\captionof{table}{Shared Layer-3 checkpoints for all molecular-optimization subtasks. There is no Type-II benchmark-state path matching; Layer~3 averages these five oracle-verifiable state checks.}
\label{tab:app_l3_molopt}
\end{table*}

The fine-grained molecule-editing and molecular-understanding tables are omitted here because their per-task values are already reported in the main text.
The remaining appendix tables focus on implementation-level reaction-prediction and molecular-optimization subtasks, where the main text reports grouped task results.

\subsection{Fine-Grained Evaluation on Reaction-Prediction Subtasks}
\label{sec:app_fine_rxnpred}

Reaction prediction has 11 active implementation subtasks, so the grouped table in the main text hides substantial variation.
Tables~\ref{tab:app_rxnpred_l1_fine} and~\ref{tab:app_rxnpred_l3_fine} report all model values at the implementation-subtask level.
The key pattern remains the same as in the main text: some subtasks have reasonable outcome accuracy, but Type-II benchmark-state agreement is much lower.
For example, condition ranking reaches 1.000 Type-I all-pass for GPT-5.2 and Claude-Sonnet, but the best Type-II agreement is only 0.135.
Reagent recommendation, solvent recommendation, yield prediction, and elementary-step prediction all have zero Type-II agreement for every model under the current all-field criterion.

\begin{table*}[htbp]
\centering
\small
\setlength{\tabcolsep}{2.6pt}
\resizebox{\textwidth}{!}{
\begin{tabular}{l rrrrrrrrrrr}
\hline
\textbf{Model} & \textbf{Major} & \textbf{Retro} & \textbf{NEP} & \textbf{Byprod.} & \textbf{Cond.} & \textbf{Cat.} & \textbf{Reag.} & \textbf{Solv.} & \textbf{Yield} & \textbf{Template} & \textbf{Mech.} \\
\hline
Qwen3.5+ & 56.5 & 18.5 & 44.0 & \textbf{35.5} & 35.5 & 40.0 & 33.0 & 22.5 & 29.5 & 90.0 & 51.5 \\
DeepSeek-V4 & 38.5 & 15.0 & 34.0 & 27.5 & 34.0 & 43.0 & \textbf{36.0} & 26.5 & 28.8 & 88.0 & 64.0 \\
GPT-5.2 & 28.5 & 8.0 & 20.5 & 23.0 & 33.0 & 32.5 & 34.0 & 17.0 & 24.1 & 91.0 & 54.5 \\
Gemini-3.1 & \textbf{70.0} & \textbf{28.5} & \textbf{55.0} & 33.0 & 37.0 & \textbf{57.0} & 32.5 & 28.0 & 29.9 & \textbf{95.5} & \textbf{77.0} \\
DeepSeek-V3.2 & 15.5 & 2.0 & 17.0 & 22.5 & 34.5 & 27.5 & 25.5 & 21.0 & 30.3 & 88.5 & 47.5 \\
Doubao-2Pro & 50.5 & 12.5 & 35.5 & 33.5 & 33.5 & 40.0 & 32.0 & \textbf{29.0} & 29.5 & 92.0 & 64.0 \\
GLM-5.1 & 52.5 & 8.0 & 28.5 & 26.0 & \textbf{38.0} & 43.5 & 33.5 & 24.0 & 30.9 & 93.5 & 68.5 \\
Claude-Sonnet & 56.5 & 20.5 & 40.0 & 29.5 & 36.5 & 48.5 & 33.5 & 23.0 & \textbf{23.3} & 95.0 & 73.0 \\
\hline
\end{tabular}}
\caption{Fine-grained reaction-prediction Layer~1 results. Values are top-1 accuracy percentages except Yield, which reports MAE (lower is better). Best values within each subtask are bolded.}
\label{tab:app_rxnpred_l1_fine}
\end{table*}

\begin{table*}[!htbp]
\centering
\small
\setlength{\tabcolsep}{2.6pt}
\resizebox{\textwidth}{!}{
\begin{tabular}{l rrrrrrrrrrr rrrrrrrrrrr}
\hline
& \multicolumn{11}{c}{\textbf{Type-I all-pass}} & \multicolumn{11}{c}{\textbf{Type-II all-match}} \\
\cline{2-12}\cline{13-23}
\textbf{Model} & \textbf{Major} & \textbf{Retro} & \textbf{NEP} & \textbf{Byprod.} & \textbf{Cond.} & \textbf{Cat.} & \textbf{Reag.} & \textbf{Solv.} & \textbf{Yield} & \textbf{Temp.} & \textbf{Mech.} & \textbf{Major} & \textbf{Retro} & \textbf{NEP} & \textbf{Byprod.} & \textbf{Cond.} & \textbf{Cat.} & \textbf{Reag.} & \textbf{Solv.} & \textbf{Yield} & \textbf{Temp.} & \textbf{Mech.} \\
\hline
Qwen3.5+ & 0.425 & 0.140 & 0.440 & 0.380 & 0.995 & 0.400 & 0.315 & 0.225 & 0.000 & 0.710 & 0.495 & 0.425 & 0.185 & 0.000 & \textbf{0.335} & 0.110 & 0.400 & 0.000 & 0.000 & 0.000 & 0.710 & 0.495 \\
DeepSeek-V4 & 0.260 & 0.095 & 0.290 & 0.205 & 0.985 & 0.415 & 0.335 & 0.265 & 0.015 & 0.590 & 0.585 & 0.210 & 0.120 & 0.000 & 0.200 & 0.120 & 0.430 & 0.000 & 0.000 & 0.000 & 0.560 & 0.435 \\
GPT-5.2 & 0.220 & 0.065 & 0.205 & 0.355 & \textbf{1.000} & 0.325 & \textbf{0.340} & 0.170 & 0.015 & 0.720 & 0.530 & 0.225 & 0.105 & 0.000 & 0.255 & 0.080 & 0.335 & 0.000 & 0.000 & 0.000 & 0.730 & 0.535 \\
Gemini-3.1 & \textbf{0.495} & \textbf{0.215} & \textbf{0.550} & 0.380 & 0.995 & \textbf{0.570} & 0.315 & 0.280 & 0.015 & 0.760 & \textbf{0.750} & \textbf{0.500} & \textbf{0.280} & 0.000 & 0.330 & 0.125 & \textbf{0.570} & 0.000 & 0.000 & 0.000 & 0.765 & \textbf{0.755} \\
DeepSeek-V3.2 & 0.085 & 0.005 & 0.160 & 0.115 & 0.995 & 0.250 & 0.250 & 0.200 & 0.010 & 0.585 & 0.445 & 0.095 & 0.055 & 0.000 & 0.110 & \textbf{0.135} & 0.300 & 0.000 & 0.000 & 0.000 & 0.590 & 0.460 \\
Doubao-2Pro & 0.330 & 0.075 & 0.310 & 0.195 & 0.995 & 0.380 & 0.320 & \textbf{0.290} & \textbf{0.020} & 0.605 & 0.480 & 0.340 & 0.125 & 0.000 & 0.205 & 0.120 & 0.400 & 0.000 & 0.000 & 0.000 & 0.645 & 0.505 \\
GLM-5.1 & 0.380 & 0.060 & 0.285 & 0.255 & 0.990 & 0.435 & 0.325 & 0.225 & 0.010 & 0.600 & 0.435 & 0.390 & 0.110 & 0.000 & 0.220 & 0.125 & 0.435 & 0.000 & 0.000 & 0.000 & 0.605 & 0.455 \\
Claude-Sonnet & 0.450 & 0.150 & 0.395 & \textbf{0.395} & \textbf{1.000} & 0.485 & 0.325 & 0.230 & 0.010 & \textbf{0.825} & 0.720 & 0.475 & 0.210 & 0.000 & 0.285 & 0.115 & 0.485 & 0.000 & 0.000 & 0.000 & \textbf{0.830} & 0.725 \\
\hline
\end{tabular}}
\caption{Fine-grained reaction-prediction Layer~3 results. The left block reports Type-I all-pass, and the right block reports Type-II all-match. Columns with all-zero Type-II results are left unbolded. ``Temp.'' denotes reaction-template selection.}
\label{tab:app_rxnpred_l3_fine}
\end{table*}

\subsection{Fine-Grained Evaluation on Molecular-Optimization Subtasks}
\label{sec:app_fine_molopt}

Molecular optimization also benefits from separating outcome and process tables.
Table~\ref{tab:app_molopt_l1_fine} reports success rates for all single- and dual-objective subtasks, while Table~\ref{tab:app_molopt_l3_fine} reports the oracle-verified optimization state score.
The outcome table shows that single-objective physicochemical tasks are comparatively easy, but coupled objectives remain difficult: LogP+solubility peaks at only 2.0 dual-success rate and GSK3$\beta$+LogP peaks at 6.0.
The process table shows that some dual-objective traces still obtain moderate oracle-verified state scores, which supports the conclusion that plausible edit rationales do not necessarily satisfy the coupled property objectives.

\begin{table*}[!ht]
\centering
\small
\setlength{\tabcolsep}{2.4pt}
\resizebox{\textwidth}{!}{
\begin{tabular}{l rrrrrr rrrrrr}
\hline
\textbf{Model} & \textbf{DRD2} & \textbf{GSK3b} & \textbf{JNK3} & \textbf{LogP} & \textbf{QED} & \textbf{Sol.} & \textbf{DRD2+LogP} & \textbf{DRD2+Sol.} & \textbf{GSK3b+LogP} & \textbf{LogP+QED} & \textbf{LogP+Sol.} & \textbf{QED+Sol.} \\
\hline
Qwen3.5+ & 50.0 & 30.0 & 30.0 & 91.7 & 76.7 & 92.5 & 2.0 & 4.0 & 2.0 & 14.0 & 0.0 & 6.0 \\
DeepSeek-V4 & 50.0 & 30.0 & 30.0 & 50.0 & 64.2 & 85.8 & 10.0 & 2.0 & 2.0 & 20.0 & 0.0 & 16.0 \\
GPT-5.2 & 65.0 & 35.0 & 39.2 & 85.8 & 79.2 & 86.7 & 10.0 & 8.0 & 0.0 & 12.0 & 0.0 & 10.0 \\
Gemini-3.1 & 70.0 & 45.8 & 38.3 & \textbf{92.5} & 87.5 & \textbf{99.2} & \textbf{16.0} & \textbf{12.0} & 2.0 & \textbf{22.0} & \textbf{2.0} & 14.0 \\
DeepSeek-V3.2 & 42.5 & 34.2 & 35.8 & 74.2 & 75.0 & 80.8 & 10.0 & 4.0 & 0.0 & 10.0 & 0.0 & 12.0 \\
Doubao-2Pro & 56.7 & 40.0 & 40.8 & 89.2 & 73.3 & 85.0 & 12.0 & 4.0 & 0.0 & 6.0 & 0.0 & 2.0 \\
GLM-5.1 & 65.8 & 40.8 & 36.7 & 85.8 & 87.5 & 85.0 & 8.0 & 6.0 & 2.0 & 20.0 & \textbf{2.0} & 18.0 \\
Claude-Sonnet & \textbf{71.7} & \textbf{57.5} & \textbf{49.2} & 85.8 & \textbf{96.7} & 93.3 & 12.0 & \textbf{12.0} & \textbf{6.0} & \textbf{22.0} & 0.0 & \textbf{26.0} \\
\hline
\end{tabular}}
\caption{Fine-grained molecular-optimization Layer~1 results. Single-objective subtasks use success rate; dual-objective subtasks use dual-success rate. Best values within each subtask are bolded.}
\label{tab:app_molopt_l1_fine}
\end{table*}

\begin{table*}[!ht]
\centering
\small
\setlength{\tabcolsep}{2.4pt}
\resizebox{\textwidth}{!}{
\begin{tabular}{l rrrrrr rrrrrr}
\hline
\textbf{Model} & \textbf{DRD2} & \textbf{GSK3b} & \textbf{JNK3} & \textbf{LogP} & \textbf{QED} & \textbf{Sol.} & \textbf{DRD2+LogP} & \textbf{DRD2+Sol.} & \textbf{GSK3b+LogP} & \textbf{LogP+QED} & \textbf{LogP+Sol.} & \textbf{QED+Sol.} \\
\hline
Qwen3.5+ & 0.471 & 0.458 & 0.462 & 0.456 & 0.529 & 0.469 & 0.535 & 0.511 & 0.497 & 0.640 & 0.458 & 0.514 \\
DeepSeek-V4 & 0.481 & 0.470 & 0.486 & 0.440 & 0.507 & 0.462 & 0.554 & 0.544 & 0.516 & 0.638 & 0.484 & 0.519 \\
GPT-5.2 & 0.453 & 0.474 & 0.433 & 0.469 & 0.514 & 0.477 & 0.511 & 0.516 & 0.502 & 0.577 & 0.450 & 0.532 \\
Gemini-3.1 & 0.546 & 0.506 & \textbf{0.537} & \textbf{0.528} & \textbf{0.585} & 0.536 & \textbf{0.583} & 0.595 & \textbf{0.568} & \textbf{0.664} & \textbf{0.519} & \textbf{0.593} \\
DeepSeek-V3.2 & 0.460 & 0.447 & 0.420 & 0.437 & 0.493 & 0.447 & 0.511 & 0.471 & 0.481 & 0.569 & 0.419 & 0.482 \\
Doubao-2Pro & 0.484 & 0.457 & 0.479 & 0.497 & 0.559 & 0.489 & 0.542 & 0.509 & 0.510 & 0.598 & 0.518 & 0.536 \\
GLM-5.1 & 0.528 & \textbf{0.511} & 0.519 & 0.465 & 0.580 & 0.429 & 0.542 & 0.534 & 0.543 & 0.654 & 0.462 & 0.559 \\
Claude-Sonnet & \textbf{0.594} & 0.504 & 0.512 & 0.493 & 0.577 & \textbf{0.548} & 0.560 & \textbf{0.628} & 0.484 & 0.621 & 0.513 & 0.576 \\
\hline
\end{tabular}}
\caption{Fine-grained molecular-optimization Layer~3 results. Values are average oracle-verified optimization state scores. Best values within each subtask are bolded.}
\label{tab:app_molopt_l3_fine}
\end{table*}

\section{Prompt Templates and System Instructions}
\label{sec:app_prompts}

This section documents the representative prompt templates used in the benchmark pipeline.
The implementation contains task-specific variants, but all variants share the same three prompt roles: molecule-edit instruction generation, GT-conditioned candidate-reference construction, and final evaluation without GT.
For reproducibility, the prompts below preserve the required system-level constraints, input fields, output format, and step names.
The concrete scripts instantiate the placeholders with the corresponding sample fields and task-specific objective names.

\subsection{Molecule-Editing Instruction Generation}
\label{sec:app_prompt_instruction_generation}

Molecule-editing instructions are generated only during dataset construction.
The generator receives the reaction-derived source-target pair and structural-change metadata, and outputs a short, site-specific edit instruction.
In our construction pipeline, this step used GPT-5.4.

\begin{lstlisting}[style=promptbox,caption={Prompt template for molecule-editing instruction generation.},label={lst:prompt_moledit_instruction}]
SYSTEM:
You are an expert organic chemist and molecular editor. Given a source molecule, a target molecule, and reaction-derived structural-change metadata, write one concise natural-language instruction that asks a model to transform the source molecule into the target molecule.

Rules:
1. Describe the molecular edit, not the reaction conditions.
2. Mention the local site when it is chemically identifiable, e.g., "the aryl bromide", "the Boc-protected piperazine nitrogen", or "the carboxylic acid group".
3. Use one of three edit semantics: add, delete, or substitute.
4. Do not reveal the target SMILES.
5. Do not mention reagents unless the reagent name is the clearest way to identify the incoming fragment.
6. Return JSON only.

USER:
Source SMILES: {source_smiles}
Target SMILES: {target_smiles}
Reaction class: {reaction_class}
Edit type: {edit_type}
Changed atoms/fragments: {structural_change_metadata}
Source-target similarity: {tanimoto}
Heavy-atom difference: {heavy_atom_delta}

Return exactly:
{
  "instruction": "<one sentence edit instruction>",
  "site": "<short site description>",
  "edit_type": "add|delete|substitute",
  "confidence": <0.0-1.0>
}
\end{lstlisting}

\subsection{GT-Injected Reference-Trace Construction}
\label{sec:app_prompt_gt_injection}

Reference traces are built with GT injection: the model is given the final answer or calibrated label, but must still fill the same formal reasoning template used later for evaluation.
These prompts are used only to construct verified step-level references; evaluated models never receive the GT fields.

\begin{lstlisting}[style=promptbox,caption={Representative GT-conditioned candidate-reference prompt for molecule editing, shown for substitution.},label={lst:prompt_gt_moledit}]
SYSTEM:
You are an expert computational chemist. Perform a molecular SUBSTITUTION edit and produce a fully verified reasoning chain.

Unified step format:
Step 1 [ANCHOR_IDENTIFICATION]: identify the substitution center, the removed group, and the incoming fragment.
  FORMAL: INDEXED_SMILES + INSTRUCTION --> ANCHOR(idx=<n>, element="<X>") + REMOVE_GROUP(smiles="<old>") + ADD_FRAGMENT(smiles="<new>")
Step 2 [REMOVE_GROUP_SIZE]: count heavy atoms in REMOVE_GROUP.
  FORMAL: REMOVE_GROUP(smiles="<old>") --> REMOVE_HEAVY(<k_old>)
Step 3 [ADD_FRAGMENT_SIZE]: count heavy atoms in ADD_FRAGMENT.
  FORMAL: ADD_FRAGMENT(smiles="<new>") --> ADD_HEAVY(<k_new>)
Step 4 [PRODUCT_CONSTRUCTION]: construct the main organic product.
  FORMAL: SMILES + ANCHOR(idx=<n>) + REMOVE_GROUP("<old>") + ADD_FRAGMENT("<new>") --> PRODUCT_SMILES("<product>")
Step 5 [HEAVY_ATOM_VERIFICATION]: verify source/product heavy-atom counts.
  FORMAL: SMILES[n_heavy=<a>] + PRODUCT_SMILES[n_heavy=<b>] --> HEAVY_ATOM_DELTA(<b-a>)
Step 6 [RING_VERIFICATION]: verify source/product ring counts.
  FORMAL: SMILES[n_rings=<c>] + PRODUCT_SMILES[n_rings=<d>] --> RING_DELTA(<d-c>)
Answer: <product_smiles>

Strict rules: output all six steps; keep each FORMAL line on one line; product and answer must be identical; no byproducts or markdown.

USER:
Source SMILES: {src_smiles}
Indexed SMILES: {indexed_smiles}
Instruction: {instruction}
Ground Truth Product SMILES: {gt_smiles}

The ground truth is provided for reference construction only. Generate the complete reasoning chain naturally and make every step consistent with the template.
\end{lstlisting}

\begin{lstlisting}[style=promptbox,caption={Representative GT-conditioned candidate-reference prompt for molecular understanding, shown for ring counting.},label={lst:prompt_gt_molund}]
SYSTEM:
You are an expert computational chemist specializing in cheminformatics and SMARTS notation. Count a specified ring type in a molecule using a fully verified formal reasoning chain.

Unified step format:
Step 1 [TARGET_SMARTS]: identify the SMARTS pattern for the target ring type.
  FORMAL: TASK("count <ring_type>") --> SMARTS("<ring_smarts>")
Step 2 [TOTAL_RINGS]: count all SSSR rings in the molecule.
  FORMAL: SMILES("<molecule>") --> RING_COUNT_TOTAL(<n_total>)
Step 3 [RING_LOCATIONS]: apply the SMARTS and enumerate all matches.
  FORMAL: SMARTS("<ring_smarts>") + SMILES("<molecule>") --> MATCH_ATOMS([<n> matches: <site_1>; ...])
Step 4 [ACCEPTED_COUNT]: count accepted target-ring matches.
  FORMAL: MATCH_ATOMS([<n> matches]) --> COUNT(<n>)
Step 5 [REJECTED_COUNT]: subtract accepted matches from total rings.
  FORMAL: COUNT(<n>) + RING_COUNT_TOTAL(<n_total>) --> REJECTED(<n_total-n>)
Answer: <n>

Strict rules: use valid SMARTS; keep arithmetic consistent; answer must equal Step 4 COUNT; no markdown or extra text.

USER:
Molecule SMILES: {smiles}
Ring type to count: {ring_name}
Ground Truth SMARTS: {gt_smarts}
Ground Truth Count: {gt_count}

The ground truth is provided only to construct a verified reference trace. Generate all five steps in the unified format.
\end{lstlisting}

\begin{lstlisting}[style=promptbox,caption={Representative GT-conditioned candidate-reference prompt for reaction prediction, shown for condition ranking.},label={lst:prompt_gt_rxnpred}]
SYSTEM:
You are an expert synthetic chemist. Rank three candidate reaction condition sets from best to worst predicted yield using a formally verifiable chain.

Allowed reaction classes: C-C Coupling; Heteroatom Alkylation and Arylation; Acylation; Functional Group Interconversion; Deprotection; Reduction; Oxidation; Aromatic Heterocycle Formation; Protection.
Allowed decision factors: catalyst, ligand, base, reagent, additive, solvent.

Unified step format:
Step 1 [RXN_CLASS]: classify the reaction.
  FORMAL: TASK("rank conditions") --> RXN_CLASS("<class>")
Step 2 [DECISION_FACTOR]: choose the single most important field.
  FORMAL: RXN_CLASS("<class>") --> DECISION_FACTOR("<field>")
Step 3 [PAIR_DIFFS]: compare all pairs 1/2, 1/3, and 2/3.
  FORMAL: CONDITIONS(["1","2","3"]) --> PAIR_DIFFS(1/2:<fields>; 1/3:<fields>; 2/3:<fields>)
Step 4 [PAIRWISE_PREFS]: derive three pairwise preferences.
  FORMAL: DECISION_FACTOR("<field>") + PAIR_DIFFS(...) --> PAIRWISE_PREFS(1>2; 1>3; 2>3)
Step 5 [RANKING]: aggregate preferences into a total order.
  FORMAL: PAIRWISE_PREFS(...) --> RANKING(["<best>","<middle>","<worst>"])
Step 6 [TOP2_SUPPORT]: justify the top-vs-second comparison.
  FORMAL: RANKING([...]) + PAIR_DIFFS(best/second:<fields>) --> TOP2_SUPPORT(WINNER="<best>", LOSER="<second>", FIELD="<field>")
Answer: ["<best>","<middle>","<worst>"]

Strict rules: compare all three pairs; preferences must be acyclic; answer must match Step 5; do not mention observed yields.

USER:
Coarse reaction class: {coarse_rxn_cls}
Ground truth ranking: {gt_ranking}
Reaction class: {rxn_cls}
Reactants: {reactants}
Product: {product}
Condition set 1: {cond_1}
Condition set 2: {cond_2}
Condition set 3: {cond_3}

Use the injected class and ranking only for reference construction. Fill the six-step trace naturally and consistently.
\end{lstlisting}

\begin{lstlisting}[style=promptbox,caption={Representative GT-conditioned candidate-reference prompt for molecular optimization, shown for LogP optimization.},label={lst:prompt_gt_molopt}]
SYSTEM:
You are an expert computational chemist specializing in medicinal chemistry and molecular property optimization. Optimize a source molecule for higher LogP while producing a formally verified reasoning chain.

Unified step format:
Step 1 [SCAFFOLD_IDENTIFICATION]: extract the Murcko scaffold of the source.
  FORMAL: SMILES("<src>") --> SCAFFOLD_SMILES("<scaffold>")
Step 2 [EDIT_PLAN]: choose one targeted functional-group edit.
  FORMAL: SMILES("<src>") --> EDIT_PLAN(remove="<fg_removed>"; add="<fg_added>")
Step 3 [PRODUCT_CONSTRUCTION]: construct the optimized molecule.
  FORMAL: SMILES("<src>") + EDIT_PLAN(remove="<fg_removed>"; add="<fg_added>") --> PREDICTED_SMILES("<new_mol>")
Step 4 [SCAFFOLD_PRESERVATION]: state whether the Murcko scaffold is preserved.
  FORMAL: SMILES("<src>") + PREDICTED_SMILES("<new_mol>") --> SCAFFOLD_PRESERVED(yes/no)
Step 5 [FG_CHANGE_VERIFICATION]: verify the claimed functional-group change.
  FORMAL: SMILES("<src>") + PREDICTED_SMILES("<new_mol>") + EDIT_PLAN(remove="<fg_removed>"; add="<fg_added>") --> FG_CHANGE_CONSISTENT(yes/no)
Answer: <new_mol>

Strict rules: predicted SMILES must be valid; answer must equal Step 3; at least one edit field is not "none"; no extra text.

USER:
Source molecule SMILES: {src_mol}
Current LogP value: {src_logp}
Ground Truth optimized SMILES: {tgt_mol}
Ground Truth improved LogP: {tgt_logp}

The ground truth is provided for reference construction only. Generate all five steps and arrive at the answer through the template.
\end{lstlisting}

\subsection{Final Evaluation Prompts}
\label{sec:app_prompt_final_eval}

At evaluation time, the system instructions keep the same formal step names and output discipline, but the user prompt removes all ground-truth fields.
The prompt builders also strip worked examples when they may leak answers, and append a strict evaluation-mode discipline block.

\begin{lstlisting}[style=promptbox,caption={Representative final-evaluation prompt for molecule editing.},label={lst:prompt_eval_moledit}]
SYSTEM:
Use the molecule-edit unified step format for the requested edit type. Output only Step 1 through the final Answer line. Every step must begin with "Step N [FIELD_NAME]:" and every FORMAL line must be indented by two spaces and stay on one line. Do not use markdown code fences or explanatory text outside the template.

USER:
Source SMILES: {src_smiles}
Indexed SMILES: {indexed_smiles}
Instruction: {instruction}

Generate the complete reasoning chain in the unified step format and output the edited molecule.
\end{lstlisting}

\begin{lstlisting}[style=promptbox,caption={Representative final-evaluation prompt for molecular understanding.},label={lst:prompt_eval_molund}]
SYSTEM:
Use the molecular-understanding unified step format for the requested subtask. Output only the formal steps and Answer. The answer must be derived from the parsed fields in the final step. No markdown, no greetings, and no text after the Answer line.

USER:
Molecule SMILES: {smiles}
Task-specific query: {query_field}

Generate the complete reasoning chain in the unified step format.
\end{lstlisting}

\begin{lstlisting}[style=promptbox,caption={Representative final-evaluation prompt for reaction prediction.},label={lst:prompt_eval_rxnpred}]
SYSTEM:
Use the reaction-prediction unified step format for the requested subtask. Output only the prescribed steps and the Answer line. Keep all FORMAL lines parseable. Do not mention ground-truth labels, observed yields, or hidden reference information.

USER:
Reaction class: {rxn_cls}
Reactants: {reactants}
Product or context: {product_or_context}
Candidate options or conditions: {options_or_conditions}

Generate the formal reasoning chain and final answer.
\end{lstlisting}

\begin{lstlisting}[style=promptbox,caption={Representative final-evaluation prompt for molecular optimization.},label={lst:prompt_eval_molopt}]
SYSTEM:
Use the molecular-optimization evaluation format. Output TWO parts in sequence.

Part A -- structured fields:
[SCAFFOLD_IDENTIFICATION]
Scaffold SMILES: <Murcko scaffold of source molecule>

[EDIT_PLAN]
FG Removed: <SMILES fragment removed, or "none">
FG Added: <SMILES fragment added, or "none">

[PRODUCT_CONSTRUCTION]
Predicted SMILES: <optimized molecule SMILES>
Answer: <same SMILES as Predicted SMILES>

[SCAFFOLD_PRESERVATION]
Scaffold Preserved: <yes/no>

[FG_CHANGE_VERIFICATION]
FG Change Consistent: <yes/no>

Part B -- formal reasoning chain:
Step 1 [SCAFFOLD_IDENTIFICATION]: explain scaffold extraction.
  FORMAL: SMILES("<src>") --> SCAFFOLD_SMILES("<scaffold>")
Step 2 [EDIT_PLAN]: explain the structural edit strategy.
  FORMAL: SMILES("<src>") --> EDIT_PLAN(remove="<fg_removed>"; add="<fg_added>")
Step 3 [PRODUCT_CONSTRUCTION]: construct the optimized molecule.
  FORMAL: SMILES("<src>") + EDIT_PLAN(remove="<fg_removed>"; add="<fg_added>") --> PREDICTED_SMILES("<pred>")
Step 4 [SCAFFOLD_PRESERVATION]: verify whether the scaffold is preserved.
  FORMAL: SMILES("<src>") + PREDICTED_SMILES("<pred>") --> SCAFFOLD_PRESERVED(yes/no)
Step 5 [FG_CHANGE_VERIFICATION]: verify whether the claimed FG changes match the SMILES diff.
  FORMAL: SMILES("<src>") + PREDICTED_SMILES("<pred>") + EDIT_PLAN(remove="<fg_removed>"; add="<fg_added>") --> FG_CHANGE_CONSISTENT(yes/no)
Answer: <pred_smiles>

Predicted SMILES must be parseable; Part A Answer must equal Part A Predicted SMILES and Part B Step 3.

USER:
Source molecule SMILES: {src_mol}
Current property values: {property_values}
Optimization objective: {objective_description}

Generate the complete structured output and formal reasoning chain.
\end{lstlisting}

\section{Case Studies for Process-Level Diagnosis}
\label{sec:app_case_studies}

The following cases illustrate failure modes that are difficult to see from aggregate tables alone.
Each example is a real evaluation record from the final framework, with long SMILES strings shortened only where the omitted context is not needed for the diagnosis.

\subsection{Type-I Failure: A Locally Plausible Edit Violates Ring Accounting}
\label{sec:app_case_type1}

In this molecule-editing example, Qwen3.5-Plus receives a substitution instruction: replace the fluorine atom on a pyridine ring with a pyrrolidin-1-yl group.
The model identifies the correct anchor and constructs a syntactically valid product, but its formal product accidentally reuses the ring index ``1'' from the larger scaffold inside the added pyrrolidine fragment.
The resulting SMILES is parseable, yet RDKit counts nine SSSR rings while the model claims seven.
Layer~3 Type-I therefore fails at the ring-verification step even before comparing against the GT trajectory.

\begin{lstlisting}[style=promptbox,caption={Type-I failure localized by deterministic ring verification.},label={lst:case_type1_ring}]
Task: molecule editing / substitute_v2
Model: Qwen3.5-Plus
Instruction: Substitute the fluorine atom on the pyridine ring with a pyrrolidin-1-yl group.

Model Step 1:
ANCHOR(idx=18, element="C") + REMOVE_GROUP(smiles="F") + ADD_FRAGMENT(smiles="N1CCCC1")

Model Step 4 product:
O=C1O[C@]2(...-c5ccc(N1CCCC1)nc5...)C2)c2ccccc21

Model Step 6 claim:
SMILES[n_rings=6] + PRODUCT_SMILES[n_rings=7] --> RING_DELTA(1)

Verifier:
RDKit source rings = 6
RDKit product rings = 9
s6_prod_rings_ok = false
Type-I all-pass = false
\end{lstlisting}

This is the intended role of Type-I checks: the surface edit is chemically plausible, but the formal trace encodes a product whose ring topology is inconsistent with the model's own verification statement.

\subsection{Type-II Benchmark-State Mismatch in a Well-Formed Trace}
\label{sec:app_case_type2}

The next example comes from reaction-condition ranking.
DeepSeek-V3.2 produces a complete and internally consistent six-step trace: it chooses a valid decision factor, compares all three pairs, produces acyclic pairwise preferences, and gives a ranking consistent with those preferences.
Thus all Type-I checks pass.
However, the reference ranking induced by the experimental yields is the reverse order.
The failure is therefore not a formatting or local-consistency problem, but a benchmark-state mismatch with the verified reference trace.

\begin{lstlisting}[style=promptbox,caption={Type-II failure in reaction-condition ranking.},label={lst:case_type2_ranking}]
Task: reaction prediction / condition_ranking
Model: DeepSeek-V3.2
Reaction: deoxyfluorination, Functional Group Interconversion
Condition 1 yield: 37.0
Condition 2 yield: 48.0
Condition 3 yield: 68.0
Ground-truth ranking: ["3", "2", "1"]

Model trace:
Step 2 DECISION_FACTOR: base
Step 3 PAIR_DIFFS: 1/2:base; 1/3:base; 2/3:base
Step 4 PAIRWISE_PREFS: 1>2; 1>3; 2>3
Step 5 RANKING: ["1", "2", "3"]
Answer: ["1", "2", "3"]

Verifier:
Type-I all-pass = true
Layer-2 State Score = 1.0
Type-II ranking match = false
Type-II all-fields match = false
\end{lstlisting}

This case shows why Layer~2 and Type-I Layer~3 are insufficient by themselves: the model follows the scientific template, but assigns the wrong chemical preference to the base series.

\section{Responsible Artifact Use and Reproducibility Details}
\label{sec:app_responsible_artifact}

This appendix summarizes artifact provenance, redistribution boundaries, and implementation settings relevant to responsible release and reproducibility.
It complements Appendix~\ref{sec:appendix}; it is not a human-subject or deployment-risk statement.

\subsection{Artifact Sources, Licenses, and Redistribution Boundaries}
\label{sec:app_artifact_sources}

\begin{center}
\scriptsize
\setlength{\tabcolsep}{3pt}
\begin{tabular}{p{0.29\columnwidth} p{0.63\columnwidth}}
\hline
\textbf{Source} & \textbf{Role, terms, and release boundary} \\
\hline
PubChem~\cite{kim2016pubchem} &
Supplemental molecule sampling; public NCBI resource with contributor-specific terms; no bulk redistribution. \\
ChEMBL 36~\cite{gaulton2017chembl} &
Bioactive molecules and matched-pair/property construction; CC BY-SA 3.0 / EBI terms; derived records only. \\
ZINC / ZINC250K~\cite{irwin2012zinc} &
Drug-like molecule pools; public ZINC access and citation terms apply; no source-database mirror. \\
Schneider 50K / USPTO~\cite{schneider2016s50k,lowe2012extraction} &
Reaction-derived molecule edits; patent-derived research corpus; release only localized edits and verifier labels. \\
USPTO / ORD pools~\cite{lowe2012extraction,kearnes2021open} &
Reaction-prediction pools; source-specific public-resource terms apply; release prompts, answers, templates, and metadata. \\
Public HTE data~\cite{dreher2009guided,ahneman2018predicting,perera2018platform} &
Condition ranking and yield labels; original-paper supplementary terms vary; release only derived condition sets and labels. \\
TDC / PyTDC~\cite{huang2021tdc} &
DRD2, GSK3$\beta$, and JNK3 oracles; TDC package and oracle terms apply; no bulk TDC redistribution. \\
RDKit~\cite{rdkit2024} and runtime packages &
Parsing, canonicalization, SMARTS, fingerprints, scaffolds, oracles, API/data utilities; open-source dependencies documented for reproduction. \\
\hline
\end{tabular}
\end{center}

We do not redistribute bulk upstream databases.
The released artifact contains anonymized derived benchmark records, task schemas, labels, prompt templates, formal reasoning templates, split metadata, and verifier descriptions where permitted.

\subsection{Intended Use}
\label{sec:app_intended_use}

The released benchmark is intended for non-commercial research evaluation of LLM chemical reasoning traces.
It is not intended to be deployed as a synthesis planner, drug-design system, laboratory recommendation tool, safety-decision system, or substitute for expert chemical review.
The released records are derived evaluation examples for benchmarking template adherence, final-answer correctness, and verifier-addressable reasoning consistency.

\subsection{Model Evaluation Setup and Computational Budget}
\label{sec:app_model_eval_setup}

\begin{center}
\scriptsize
\setlength{\tabcolsep}{3pt}
\begin{tabular}{p{0.28\columnwidth} p{0.64\columnwidth}}
\hline
\textbf{Item} & \textbf{Setting} \\
\hline
Models &
Qwen3.5+, DeepSeek-V4, DeepSeek-V3.2, Doubao-2Pro, GLM-5.1, GPT-5.2, Gemini-3.1, Claude-Sonnet. \\
Access and period &
All LLMs were API-accessed in May 2026; local execution was only for parsing, RDKit/TDC oracles, and metric computation. \\
Calls and decoding &
5,620 prompts/model; about 44,960 calls for eight models excluding retries; temperature 0.1, \texttt{max\_tokens}=32,768, timeout 800s. \\
Budget notes &
Seed 42 for sampling utilities; no local LLM training/inference; API model sizes/training data unavailable; exact cost depends on provider tokenization/pricing. \\
\hline
\end{tabular}
\end{center}

\subsection{Software Environment and Chemical Evaluation Parameters}
\label{sec:app_software_params}

\begin{center}
\scriptsize
\setlength{\tabcolsep}{3pt}
\begin{tabular}{p{0.28\columnwidth} p{0.64\columnwidth}}
\hline
\textbf{Component} & \textbf{Version / setting} \\
\hline
Environment &
Python 3.12.13; RDKit 2023.09.6; NumPy 1.26.4; pandas 2.3.3; scikit-learn 1.2.2; OpenAI client 2.28.0; requests 2.32.5; tqdm 4.67.3; PyTDC installed, version not exposed. \\
RDKit operations &
\texttt{MolFromSmiles} with default sanitization; \texttt{MolToSmiles(..., canonical=True)}; \texttt{MurckoScaffoldSmiles(..., includeChirality=False)}; \texttt{CalcNumRings}; \texttt{CalcMolFormula}. \\
Fingerprints &
Morgan radius 2; 2,048 bits for molecular-understanding Tanimoto and reaction-prediction FTS utilities; 1,024 bits for molecular-optimization molecule/scaffold similarity. \\
Oracles &
LogP: RDKit \texttt{MolLogP}; QED: RDKit \texttt{QED.qed}; solubility: $0.16 - 0.63\log P - 0.0062\,MW + 0.066\,HBD - 0.074\,HBA$; DRD2/GSK3$\beta$/JNK3: \texttt{TDC.Oracle(name=...)}. \\
Randomness &
Default seed 42 in released evaluation and split/sampling utilities. \\
\hline
\end{tabular}
\end{center}

\subsection{Reproducibility Package Organization}
\label{sec:app_release_package}

The anonymized supplement contains \texttt{anonymous\_data/} and \texttt{anonymous\_software/}: task schemas, formal templates, prompt templates, active split metadata, sample examples, verifier rule descriptions, one-to-one aligned raw/process-evaluation records, and the generation, parsing, verification, oracle-wrapper, aggregation, validation, and API-facing evaluation utilities needed to reproduce the released framework.